\pdfoutput=1

\documentclass[11pt]{article}

\usepackage[]{ACL2023}

\usepackage{times}
\usepackage{latexsym}

\usepackage[T1]{fontenc}

\usepackage[utf8]{inputenc}

\usepackage{microtype}

\usepackage{inconsolata}

\usepackage{amsmath}
\usepackage{amsfonts}
\newtheorem{theorem}{Theorem}
\newtheorem{lemma}{Lemma}

\usepackage{multirow}
\usepackage{mathrsfs}
\usepackage{graphics}
\usepackage{subfigure}
\usepackage{wrapfig}
\usepackage{epstopdf}
\usepackage{pdfpages}
\usepackage{diagbox}
\usepackage{ulem}
\usepackage{MnSymbol}

\usepackage{tabularx}
\usepackage{diagbox}
\usepackage{makecell}

\title{MGR: Multi-generator Based Rationalization}
\author{
Wei Liu$^1$ \quad Haozhao Wang$^1$\thanks{\ \   Corresponding author} \quad  Jun Wang$^2$\thanks{\ \ This paper is a collaborative work between the Intelligent and Distributed Computing Laboratory at Huazhong University of Science and Technology, and \href{https://www.iwudao.tech/}{iWudao Tech}.}\quad Ruixuan Li$^1$ \\ \textbf{Xinyang Li}$^1$ \quad \textbf{Yuankai Zhang}$^1$ \quad \textbf{Yang Qiu}$^1$
\\$^1$School of Computer Science and Technology, \\
$^1$Huazhong University of Science and Technology, Wuhan City, Hubei Province, China\\  $^2$iWudao Tech\\
$^1$\texttt{\{idc\_lw, hz\_wang, rxli, lxy722, yuankai\_zhang, anders\}@hust.edu.cn } \\ $^2$\texttt{jwang@iwudao.tech}\\
}

\begin{document}

\normalem
\maketitle
\begin{abstract}
Rationalization is to employ a generator and a predictor to construct a self-explaining NLP model in which the generator selects a subset of human-intelligible pieces of the input text to the following predictor.
However, rationalization suffers from two key challenges, i.e., spurious correlation and degeneration, where the predictor overfits the spurious or meaningless pieces solely selected by the not-yet well-trained generator and in turn deteriorates the generator. 
Although many studies have been proposed to address the two challenges, they are usually designed separately and do not take both of them into account.
In this paper, we propose a simple yet effective method named MGR to simultaneously solve the two problems. The key idea of MGR is to employ multiple generators such that the occurrence stability of real pieces is improved and more meaningful pieces are delivered to the predictor.
Empirically\footnote{\href{https://github.com/jugechengzi/Rationalization-MGR}{https://github.com/jugechengzi/Rationalization-MGR}}, we show that MGR improves the F1 score by up to $20.9\%$ as compared to state-of-the-art methods.
\end{abstract}

\section{Introduction}\label{sec:introduction}

\begin{figure}[t]
    \centering
    \includegraphics[width=0.99\columnwidth]{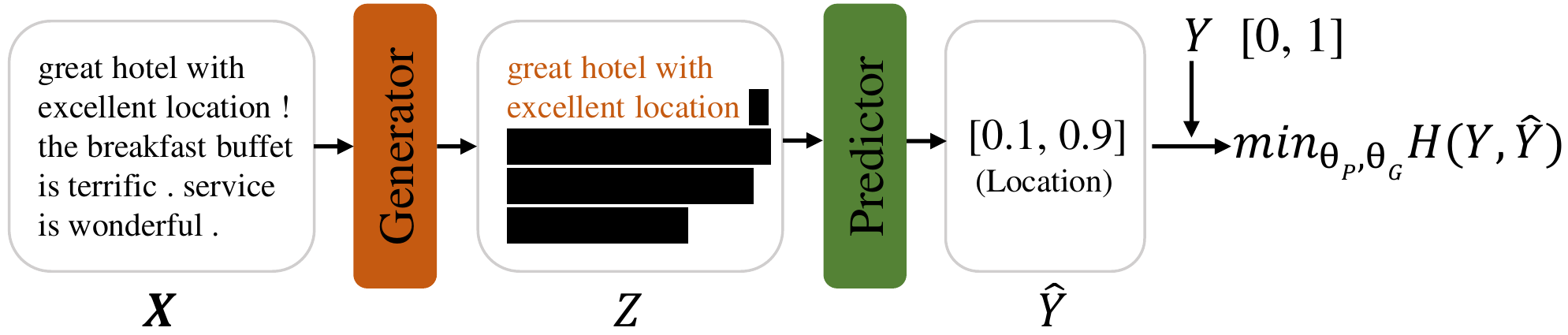}
    \caption{The standard rationalization framework RNP. $X,Z,\hat{Y},Y$ represent the input text, rationale, prediction and the groundtruth label, respectively.   }
    \label{fig:RNP}
\end{figure}

The widespread use of deep learning in NLP models has led to increased concerns about interpretability. To solve this problem, \citet{rnp} proposed rationalization framework RNP in which a generator selects human-intelligible subsets (i.e., rationales) from the input text and feeds them to the subsequent predictor that maximizes the text classification accuracy, as shown in Figure~\ref{fig:RNP}. 
Unlike post-hoc approaches for explaining black-box models, the RNP framework has the built-in self-explaining ability through a cooperative game between the generator and the predictor. RNP and its variants have become one of the mainstreams to facilitate the interpretability of NLP models \citep{interlocking,liufr, liudr}. Notably, given the versatility of the self-explaining rationalization framework, such methods have significant potential for application in diverse fields such as multi-aspect recommender systems \citep {deng2023multi} and computer vision \citep{GDM}. 

Despite its strength, rationalization schemes are notoriously hard to train. Two main training obstacles are the spurious correlations \cite{invarant} and the degeneration \citep{rethinking}. As shown in the example of Table~\ref{tab:example of diff}(a), the problem of spurious correlations is that the predictor mistakenly makes a correlation between the label on some specific aspect and the spurious pieces on another similar aspect, which commonly exists in multi-aspect classification \citep{invarant,counter,interventional}. Degeneration means that the predictor may overfit to meaningless rationales generated by the not yet well-trained generator \citep{rethinking}, causing the converged generator tends to select these uninformative rationales, which is illustrated in the example of Table~\ref{tab:example of diff}(b). 

Many prior efforts have separately considered the problem of spurious correlations or degeneration in rationalization. 
For instance, to solve the problem of spurious correlations, some recent methods leverage the idea of causal inference to build the causal relationship between the rationale and label \citep{invarant, interventional}. The common idea to address the degeneration problem is to introduce some auxiliary modules such that the predictor has access to the full texts, and thus it cannot overfit the meaningless rationales solely provided by the generator \citep{rethinking, dmr, interlocking}.

\begin{table*}[t]
\subtable[An example of spurious correlation]{
\begin{tabularx}{\textwidth}{X}
\hline\hline
	
    \textbf{Label (Aroma):} Positive\quad \textbf{Prediction:} Positive\\
    \textbf{Text:} the appearance was nice . dark gold with not much of a head but nice lacing when it started to dissipate . \
    {\underline{\textcolor{blue}{the}}  \textcolor{blue}{smell was ever so hoppy with a hint of the grapefruit flavor} \textcolor{blue}{that 's} \textcolor{blue}{contained}} \textcolor{blue}{within} \textcolor{blue}{.} \  \underline{the taste was interesting , up front tart grapefruit , not sweet in the} least . more like grapefruit rind even . slight hint of hops and seemingly no malt . the mouth feel was crisp , with some biting carbonation . drinkability was easily above average due to the crispness and lack of sweetness . not the usual taste you expect when drinking a fruit beer . in fact this is my favorite fruit beer ever . \\
    
    \hline
\hline
\end{tabularx}
\label{tab: spurious correlation}
}
\subtable[An example of degeneration]{
\label{tab:egdegeneration}
\begin{tabularx}{\textwidth}{X}
\hline\hline
    \textbf{Label (Appearance):} Negative \quad \textbf{Prediction:} Negative\\
    \textbf{Text:}\ { \underline{\textcolor{blue}{appearance}} \textcolor{blue}{: light yellow to} \textcolor{blue}{almost} \textcolor{blue}{clear}} smell : slight hops , but barely smelled like beer taste : little to none , like a rice lager , zero taste mouthfeel : watery and tasteless drinkability : very easy , goes down easier than water . good for drinking games.  \\
    \hline\hline
    
\end{tabularx}
}
\caption{The \textcolor{blue}{blue} piece of the text is the human-annotated rationale. Pieces of the text with \ \underline{underline}  are the rationales from RNP. (a): An example of RNP making the right sentiment prediction using the spurious correlation. If the predictor overfits the spurious correlation, it will then tell the generator to continue to select this spurious correlation as the rationale. (b): An example of RNP making the right sentiment prediction using an uninformative rationale. Initially, the generator may randomly select some uninformative candidates like “appearance” as rationales for the negative text. The predictor of RNP overfits to these uninformative rationales and classifies the sentiment according to whether “appearance” is included in the rationale. Guided by such a spoiled predictor, the generator in turn tends to select these uninformative rationales.}
\end{table*}\label{tab:example of diff}

Although these approaches may be effective at either solving the problem of spurious correlations or degeneration isolation, they are usually designed separately and do not take both of them into account. In this paper, we seek to simultaneously solve the two problems. Specifically, we identify that both two problems arise from that the predictor has only access to the limited view of pieces provided by the single generator, and thus may learn corruptly when this generator selects spurious or meaningless rationales. Besides, recent studies find that the initialization of the model has a significant impact over the training performance, which implicitly indicates that the rationalization model is hardly to train once the single generator is not well initialized \cite{jain2020faith, interlocking}.

Considering these limitations of the rationalization with one single generator, as shown in Figure~\ref{fig:mhg}, we design a novel architecture where there is a predictor but with multiple generators. These generators are initialized with different parameters. In this way, the view of the predictor is not limited to one single generator and it can have access to more meaningful rationales. We theoretically show that the occurrence stability of real rationales increases such that the predictor has lower risks at learning spurious correlations, and that the diversity of the rationales is improved such that the predictor can hardly deviate to some specific meaningless rationale. Extensive experiments conducted on three widely used rationalization benchmarks, i.e., the correlated BeerAdvocate dataset \citep{beer}, the decorrelated BeerAdvocate dataset \citep{rnp},  and the Hotel Reviews dataset \cite{hotel}, show that MGR achieves significant improvements over several state-of-the-art methods in terms of the rationale quality. 
Our contributions can be summarized as:\\
$\bullet$ To the best of our knowledge, this paper is the first to simultaneously solve the spurious correlations and degeneration problem in rationalization. We propose a simple but effective method, namely, MGR, that facilitates the predictor to have a broader view of the rationales by using multiple generators.\\
$\bullet$ We theoretically prove that using multiple generators can provide real rationales more stably such that the risk of the predictor learning spurious correlations is reduced. Besides, we prove that multiple generators can produce more diverse rationales and thus the predictor will not overfit to some specific meaningless rationale. \\
$\bullet$ We conduct extensive experiments over various datasets and show that MGR achieves an improvement by up to $20.9\%$ as compared to state-of-the-art rationalization methods in terms of F1 score.

\section{Related Work}\label{sec:related work}
\subsection{Rationalization}
The base cooperative framework of rationalization named RNP \citep{rnp} is flexible and offers a unique advantage, i.e., certification of exclusion, which means any unselected input is guaranteed to have no contribution to prediction \citep{interlocking}. However, such a method is hard to train.
To tackle this challenge, many methods have been proposed to improve RNP from different aspects. 

\noindent\textbf{Rationales sampling}. Many works focus on refining the sampling process of the rationales. \citet{2018rationalegumble} used 
Gumbel-softmax to do the reparameterization for binarized selection. \citet{hardkuma} replaced the Bernoulli sampling
distributions with rectified Kumaraswamy distributions. \citet{jain2020faith} disconnected the training
regimes of the generator and predictor via a saliency threshold. \citet{informationbottle} imposed a discrete bottleneck objective to balance the task performance and the rationale length. \citet{leakage} explored better metrics for the explanations. \citet{concept} used phrase-based concepts to conduct a self-explaining model.  These methods are orthogonal to our method.

\noindent\textbf{Degeneration}. Degeneration is one of the major problem in rationalization. To solve this problem, many efforts seek to regularize the predictor using supplementary modules which have access to the information of the full text \citep{rethinking,dmr,interlocking} such that the generator and the predictor will not collude to uninformative rationales. 3PLAYER \citep{rethinking} takes the unselected text $Z^c$ into consideration by inputting it to a supplementary predictor \emph{Predictor}$^c$. DMR \citep{dmr} tries to align the distributions of rationale with the full input text in both the output space and feature space. A2R \citep{interlocking} endows the predictor with the information of full text by introducing a soft rationale. 

\noindent\textbf{Spurious correlations}. The predictor in rationalization model may make correlations between spurious rationales and the label. To tackle this challenge, the typical methods mainly adopt causal inference to exclude the spurious correlations. For instance, \citet{invarant} introduced an environment-agnostic predictor to recognize the spurious correlations. \citet{interventional} aimed to remove the spurious correlations based on the backdoor adjustment.

\subsection{Model Ensemble Methods}
Ensemble methods that combine the outputs of several different models to improve the prediction performance and robustness have been studied for a long time \citep{bagging,stacked,boosting}. Ensemble methods train $N$ models with different datasets independently and fuse the outputs of different models during inference. This requires maintaining $N$ models and running each of them at test time, which increases the costs of computational
resources and brings obstacles to applications \citep{fusion}. 
Although our method is similar to ensemble methods to some extent, it has essential differences with ensemble methods. In our method, different generators are not trained entirely independently. In fact, they all play a cooperative game with the same predictor on one same training dataset. With different initializations, different generators can provide diverse rationales to train a robust and stable predictor at early training stage. But with the same training target and dataset, different generators can converge to get the same output \citep{gitrebasin}, thus we only need to keep one generator during inference, which is also empirically supported by the experimental results in Figure~\ref{fig:nodis and onehead}(b).

\section{Problem Definition}
\textbf{Notation}. 
$f_G(\cdot)$ and $f_P(\cdot)$ represent the generator and predictor, respectively. $\theta_G$ and $\theta_P$ represent the parameters of the generator and predictor, respectively. $\mathcal{D}$ represents the distribution of dataset.
We consider the classification problem, where the input is a text sequence  ${X}$=$[x_1,x_2,\cdots,x_l]$ with ${x}_i$ being the $i$-th token and $l$ being the number of tokens. The label of ${X}$ is a one-hot vector $Y\in\{0,1\}^c$, where $c$ is the number of categories. 

\noindent \textbf{Cooperative rationalization}. Rationalization framework consists of a generator and a predictor. The goal of the generator is to find the most informative pieces containing several tokens in the original input text $X$. 
For each sample $(X,Y)\sim \mathcal{D}$, the generator firstly outputs a sequence of binary mask $M=[m_1,\cdots,m_l]\in \{0,1\}^l$. Then, it forms the rationale $Z$ by the element-wise product of $X$ and $M$:
\begin{equation}\label{eqa:getrat}
    Z=M\odot X=[m_1x_1,\cdots,m_lx_l].
\end{equation}
To simplify the notation, we denote $Z=f_G(X)$.
In cooperative rationalization, the informativeness of the rationale $Z$ provided by the generator is measured by the negative cross entropy $-H(Y,\hat{Y}_z)$, where $\hat{Y}_z$ is the output of the predictor with the input being $Z$. Consequently, the generator and the predictor are usually optimized cooperatively:
\begin{equation}\label{eqa:objpg}    \mathop{\min}\limits_{\theta_G,\theta_P}\sum_{(X,Y) \sim \mathcal{D}}H(Y,f_P(f_G(X))).
\end{equation}

\noindent \textbf{Regularizer of shortness and coherence}.
To make the selected rationales human-intelligible, the original RNP constrains the rationales by short and coherent regularization terms. In this paper, we use the constraints updated by \citet{car}:
\begin{equation}\label{eqa:regular}
\Omega (M) = \lambda_1 | \frac{||M||_1}{l}-s| +\lambda_2\sum_{t=2}^{l} \big|m_t-m_{t-1} \big|. 
\end{equation} The first term encourages that the percentage of the tokens being selected as rationales is close to a pre-defined level $s$. The second term encourages the rationales to be coherent.

\begin{figure}
    \centering
    \includegraphics[width=0.95\columnwidth]{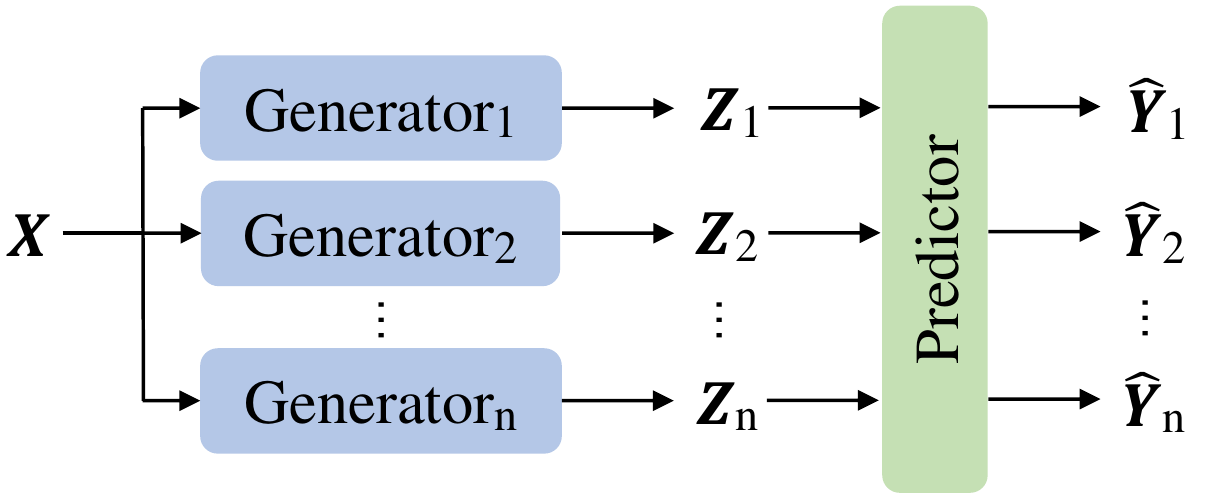}
    \caption{The architecture of MGR. $X$ is the input full text. $Z_i$ is the selected rationale and ${\hat{Y_i}}$ is the predictor's output corresponding to $Z_i$. 
    }
    \label{fig:mhg}
\end{figure}

\section{Multi-generator based Rationalization}\label{sec:method}
\subsection{Methodology}
Based on the framework of RNP, MGR uses multiple generators with different initialized parameters to help the predictor learn from diverse rationale candidates, as shown in Figure~\ref{fig:mhg}. For the convenience of comparing with previous methods in experiments, we adopt the bidirectional gated recurrent units (GRUs) \citep{gru} as the encoder which has been adopted by most previous works \citep{invarant,dmr,interlocking}.

\noindent \textbf{Training of MGR}.
MGR is to leverage each generator $f_{G_i}(\cdot)$ to process the input text $X$ in isolation, and then send the obtained rationales $Z_i$ to the predictor $f_{P}(\cdot)$ to obtain the prediction $\hat{Y_i}$. Based on Equation~\ref{eqa:objpg}, MGR computes its loss by calculating the sum of the cross entropy between $Y$ and $\hat{Y}_i$ in each generator:
\begin{align}
\begin{split}
&\mathop{\min}\limits_{\theta_{G_1},\ldots,\theta_{G_n}, \theta_P}\sum_{(X,Y)\sim \mathcal{D}} H(Y,\hat{Y}) \\
    &=\sum_{(X,Y)\sim \mathcal{D}}\sum_{i=1}^n H(Y,f_{P}(f_{G_i}(X))).
\end{split}
\end{align}
 
\noindent \textbf{Inference of MGR}. During inference, MGR only uses one generator, e.g., the first generator $Z_1$ and $\hat{Y_1}$, to provide the rationale. It's worth noting that our idea is similar to but not the same as ensemble models. We keep a set of \emph{generators} during training to help train a good predictor which in turn will promise the cooperative generator to be good. Nevertheless, we only keep the first \emph{generator} during inference (see experimental results in Figure~\ref{fig:nodis and onehead}(b)), which is efficient in terms of both time and resource consumption.

\subsection{Dealing with spurious correlations}
In this section, we seek to show that the proposed method can solve the problem of spurious correlations. Specifically, we specify the principles of our method on the case with simplified settings.

\noindent \textbf{Settings}. Similar to \citet{interlocking},  we consider the input text $X$ consists of two subsets of the features $X_1$ and $X_2$, where $X_1$ is the causal rationale belonging to the target aspect label, while $X_2$ is the comments about other aspects that correlates with the target aspect. We denote the parameters of the predictor by $\theta_P^1$ when it fits $X_1$, and by $\theta_P^0$ when it fits $X_2$. 
We also assume a payoff table as shown in Table~\ref{tab:payoff}, where we denote the negative loss as the payoff. The higher payoff indicates that the rationalization achieves better convergence results. Considering that rationalization is a cooperative game, the payoff of the generator is the same as that of the predictor. Here, $a$ can be seen as the negative loss when the predictor fits the generator's selection and $b$ is the negative loss when the predictor fits the unselected part. Since the loss is usually smaller when $f_P(\cdot)$ fits what $f_{G_i}(\cdot)$ selects, we consider $a>b$. We denote that there are $k$ \emph{generators} that select the causal rationale $X_1$ and the other $n-k$ \emph{generators} select  $X_2$. Without losing generality, we consider that the predictor is randomly initialized by interpolation between $\theta_P^0, \theta_P^1$ with $\theta_P^{\alpha}$, where $\alpha$ is a random variable in [0,1]. Similar to \citet{interlocking}\footnote{Our $\alpha$ is similar to $\pi$ in \cite{interlocking}. In \cite{interlocking}, $\pi$ is the degree that the generator tends to select $X_1$. }, we consider $\alpha$ to be the probability (or degree) that the predictor tends to fit $X_1$. If $\alpha=1$, the predictor always fits $X_1$, and if $\alpha=0$, the predictor always fits $X_2$. This is where the third column in Table~\ref{tab:payoff} comes from.   

\begin{table}[t]
    \centering
    \resizebox{0.99\columnwidth}{!}{
    \begin{tabular}{c|c|c|c}
    \hline
      \diagbox{$f_{G_i}(\cdot)$}{$f_P(\cdot)$}   & $\theta_P^1$ &$\theta_P^\alpha$&$\theta_P^0$ \\
      \hline
         Select $X_1$& $(a,a)$ &($\alpha\cdot a+(1-\alpha)\cdot b$,\ $\alpha\cdot a+(1-\alpha)\cdot b$)& $(b,b)$\\
         \hline
         Select $X_2$ & $(b,b)$&$((1-\alpha)\cdot a+ \alpha\cdot b,\ (1-\alpha)\cdot a+ \alpha\cdot b)$ & $(a,a)$\\
         \hline
    \end{tabular}
    }
    \caption{The payoff (negative loss) table of the cooperative game between the generator and the predictor. Here $a>b$, which indicates that if $f_P(\cdot)$ fits what $f_{G_i}(\cdot)$ selects, both  $f_P(\cdot)$ and $f_{G_i}(\cdot)$ get a higher payoff.}
    \label{tab:payoff}
\end{table}

\noindent\textbf{Theoretical analysis}. For the situation that $k$ \emph{generators} select $X_1$ and the predictor is $\theta_P^{\alpha}$, the expected payoff of the predictor is
\begin{equation}\label{eqa:payoff predictor}
\begin{aligned}
 R_{P}({\alpha})=&k\cdot\alpha\cdot a+k(1-\alpha)\cdot b\\
 &+(n-k)(1-\alpha)a+(n-k)\cdot\alpha\cdot b.  
 \end{aligned}
\end{equation}

With the generator's random initialization, $k$ follows a binomial distribution $\mathcal{B}(n,P_c)$, where $P_c$ depends on the dataset (subscript $c$ stands for causality). We assume that $P_c>0.5$ because the causality appears more often than the spurious correlation when the target aspect's label appears (Please refer to Appendix~\ref{proof:pc>0.5} and Table~\ref{tab:pc} for more detailed discussion). Apparently, we have 
 \begin{equation}\label{eqa:limit error}
    \lim_{n\to\infty}p(k<n-k)=0, 
 \end{equation}
of which the detailed derivation is in Appendix~\ref{proof:limit error}.

\begin{lemma}\label{lem: pred choose x1}
    If the number of generators that select $X_1$ is more than those that select $X_2$ (i.e., $k>n-k$), the predictor will be optimized to increase $\alpha$ to get a higher payoff (i.e., $\frac{\partial R_{P}({\alpha})}{\partial\alpha}>0$). 
\end{lemma}
The proof is in Appendix~\ref{proof:lemma1}. Lemma~\ref{lem: pred choose x1} indicates that if $k>n-k$, by increasing $R_P(\alpha)$ (for lower loss), the cooperative game will guide the predictor to move towards $\theta_P^1$ (by increasing $\alpha$) and fit the right causal rationale $X_1$. In turn, such a good predictor will guide the generator to select the desired causal rationales.

We denote the probability of the predictor overfits to the spurious correlation as $p_{MGR}(spu)$. According to Lemma~\ref{lem: pred choose x1}, we have 
\begin{align}
    p_{MGR}(spu)=&p(\frac{\partial R_{P}({\alpha})}{\partial\alpha}<0)\nonumber \\
    =&p(k<n-k)\\
    =&\sum_{k=0}^{(n-1)/2}(\binom{n}{k}\cdot P_c^k\cdot(1-P_c)^{n-k}), \nonumber
\end{align}
where $\binom{n}{k}$ is the combinatorial number.
Combining Lemma~\ref{lem: pred choose x1} with Equation~\ref{eqa:limit error}, we have the following theorem: 
\begin{theorem}\label{theorem: n and spurious}
For any tolerable upper bound probability $P_s$ of the predictor overfitting to spurious correlation, if $P_c>0.5$, there always exists a constant $N$ such that 
\begin{equation}
    \forall n>N, \quad p_{MGR}(spu)<P_s.
\end{equation}
\end{theorem}
The proof is deferred to Appendix~\ref{proof: n and spurious}. Theorem~\ref{theorem: n and spurious} indicates that we can reduce the risk of spurious correlations to arbitrarily low levels by increasing the number of the generator. When $n=1$, MGR becomes the vanilla RNP and we have 
\begin{equation}
    p_{RNP}(spu)=1-P_c.
\end{equation}
It is obvious that 
\begin{equation}
    \forall n>1, \quad p_{MGR}(spu)<p_{RNP}(spu).
\end{equation}

\subsection{Dealing with Degeneration}
In this section, we consider $X,Y,Z$ as random variables rather than deterministic ones. 
The principle behind Equation~\ref{eqa:objpg} is to maximize the mutual information between $Y$ and $Z$ \citep{invarant,interlocking}:
\begin{equation}
    \mathop{max}_Z I(Y;Z)=\mathop{max}_Z (H(Y)-H(Y|Z)).
\end{equation}

Since $H(Y)$ is irrelevant to $Z$, the equation is equal to minimizing $H(Y|Z)$.
Degeneration happens because the diversity of the rationales is not taken into account. Hence, the generator of RNP may get rationale candidates with low diversity (i.e., low $H(Z)$). Under this case, the predictor may overfit to some specific patterns that are contained in the limited rationale candidates and has a high risk of occurring degeneration when the rationales are merely noises.

Next, we show that MGR with multiple generators improves the diversity of the selected rationale candidates. Specifically, by viewing the rationales of different generators as different variables, we can compute the rationale diversity of MGR as
\begin{equation}
\begin{aligned}
    H(Z_{MGR})=H(Z_1,Z_2,\cdots, Z_n).
\end{aligned}
\end{equation}

\begin{theorem}\label{theorem: higher entropy with mgr}
For $\forall i\in [1,n]$, we have 
\begin{equation}
H(Z_i)\leq H(Z_1,Z_2,\cdots, Z_n) \leq \sum_{k=1}^n H(Z_k),
\end{equation}
where the right equality holds if and only if $\forall i,j,\ Z_i \upmodels Z_j$, and the left equality holds if and only if $\forall i,j,\ Z_i = Z_j$.
\end{theorem}
The proof is in Appendix~\ref{proof:higher entropy of mgr}. Theorem~\ref{theorem: higher entropy with mgr} indicates that the diversity of MGR with multiple generators is equivalent to the case with one single generator when all generators are the same. More specifically, since RNP consists of only one generator, we have 
\begin{equation}
    H(Z_{RNP})=H(Z_i)\leq \mathop{\max}_i H(Z_i),
\end{equation}
where $i \in [1,n]$. We always have $H(Z_{MGR})\geq H(Z_{RNP})$ no matter how different generators are coupled together, thus alleviating degeneration. 

Besides, Theorem~\ref{theorem: higher entropy with mgr} also indicates that the diversity of MGR achieves the maxima when all the generators are completely independent.
Accordingly, we seek to decrease the correlation between different generators to make that $H(Z_{MGR})$ gets closer to $\sum_{k=1}^n H(Z_k)$ during training which is specified in the next section. 

\subsection{Diverse Training with Separate Learning Rates}\label{sec:separate lr}

To facilitate the improvement of the diversity of rationales while guaranteeing the convergence of rationalization models, we consider that training MGR has to satisfy two conditions. First, to deal with degeneration, generators should be different from each other to guarantee that the predictor continuously learns from diverse rationales before it learns adequate information. Second, different generators should be able to achieve the same convergence result, i.e., selecting the same rationales for any given text, after the predictor has learned enough information and converged. Only in this way can we keep one single generator during inference to guarantee that MGR is efficient in terms of latency and resource consumption.

\begin{figure}[t]
    \subfigure[]{
    \label{fig:rnp}
        \includegraphics[height=2.71cm]{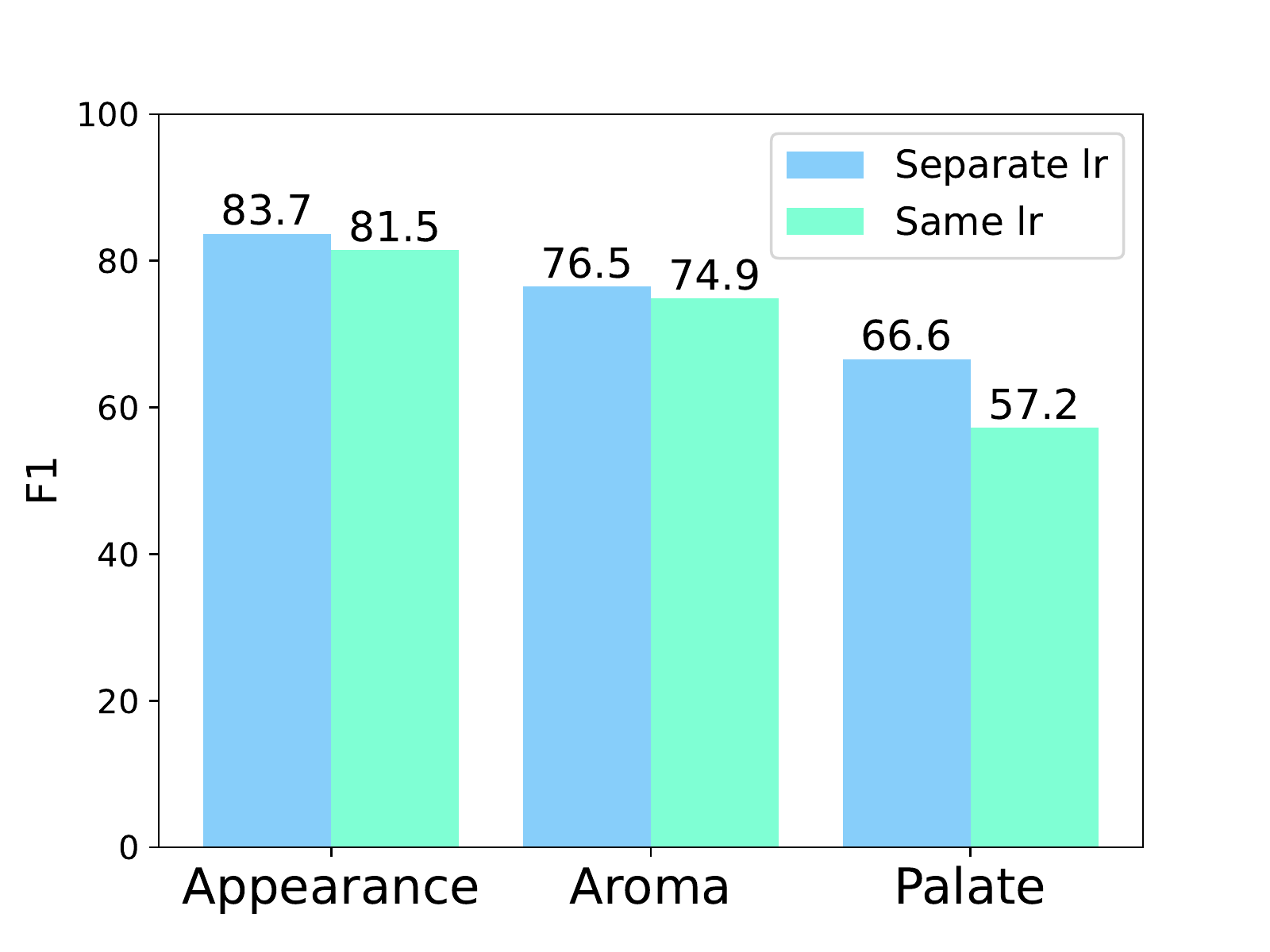}  
    }
    \subfigure[]{
    \label{fig:3player}
        \includegraphics[height=2.71cm]{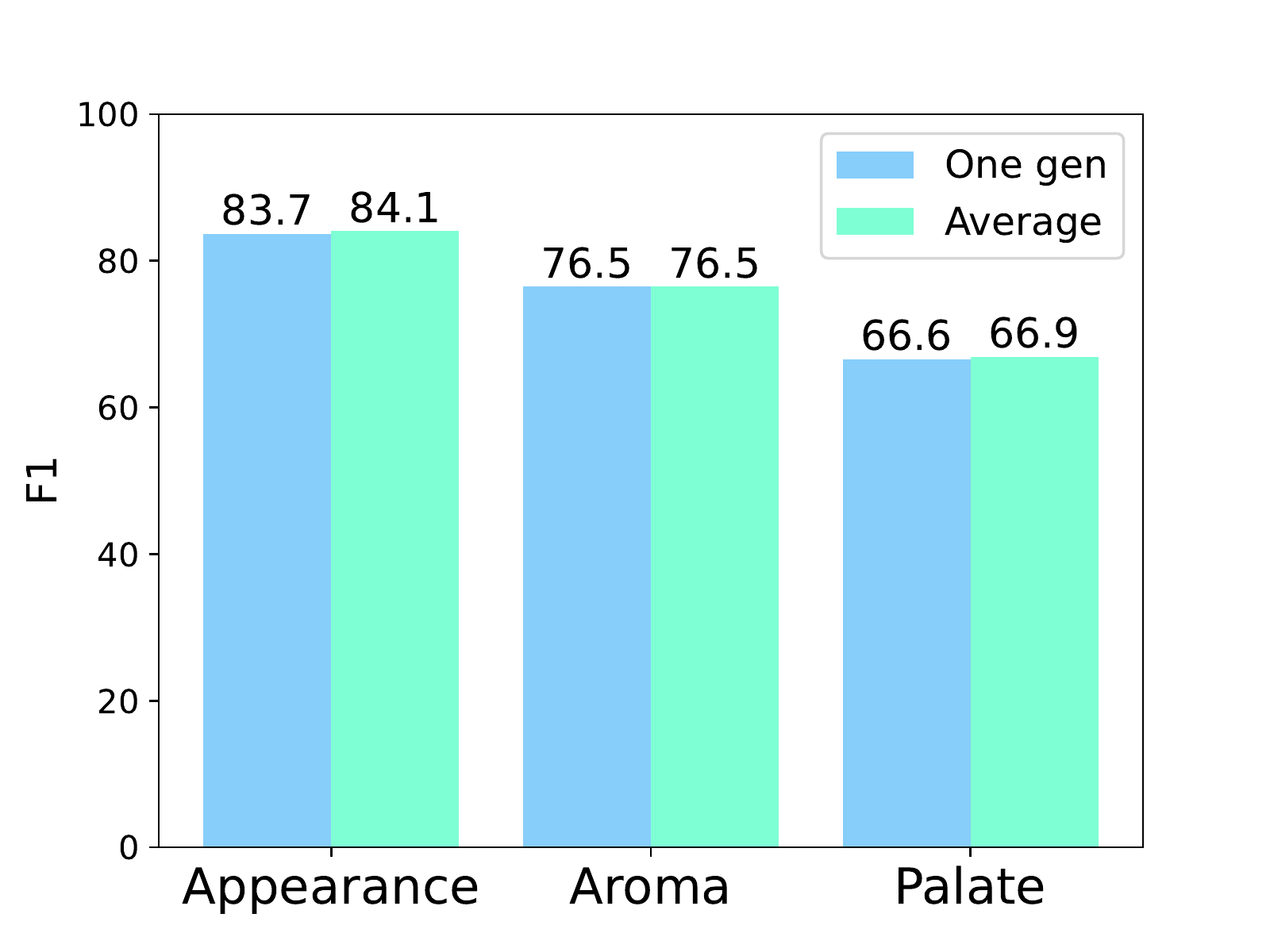}
    }
   
  \caption{Experiments on \emph{decorrelated BeerAdvocate} verifying (a) the effectiveness of separate learning rates and (b) the feasibility of keeping only the first generator.
}
  \label{fig:nodis and onehead}
\end{figure}

To satisfy the two properties, we propose separately setting the learning rates of different generators. Intuitively, separate learning rates provide different generators with different learning states in any training moment, thus keeping them diverse during the learning process. On the other side, learning rates do not modify the loss landscape of generators and thus these generators can eventually achieve the same convergence result although maybe at different speeds \cite{gitrebasin}. The argument is also empirically supported by the results in Figure~\ref{fig:nodis and onehead}(b).

Formally, we denote the learning rate of the $i$-th generator as $\eta_i$ and the loss as $\mathcal{L}$.  \emph{generator$_i$} and \emph{generator$_j$} are updated during training as:
\begin{equation}\label{eqa:different lr}
\begin{split}
    \theta_{G_i}{^\prime}&=\theta_{G_i}-\eta_i\cdot {\nabla}_{\theta_{G_i}}\mathcal{L}, \\
    \theta_{G_j}{^\prime}&=\theta_{G_j}-\eta_j{\cdot} \nabla_{\theta_{G_j}}\mathcal{L}, 
\end{split}
\end{equation}
In practice, we first find a learning rate $\eta$ and set the $i$-th generator's learning rate simply to be $i\cdot\eta$. And to alleviate the problem that the loss function of the predictor is too large due to the superposition of multiple cross-entropies from different generators, we set the learning rate of the predictor to be $\frac{\eta}{n}$.

\begin{table*}[t]
    \centering
    \resizebox{1.99\columnwidth}{!}{
    \begin{tabular}{c c c |c c| c c |c|c c| c c |c |c c| c c |c }
\hline
\multicolumn{3}{c|}{\multirow{2}{*}{Methods}} & \multicolumn{5}{c|}{Appearance} & \multicolumn{5}{c|}{Aroma} & \multicolumn{5}{c}{Palate}\\
\cline{4-18}
\multicolumn{3}{c|}{} &S& Acc & P & R &\multicolumn{1}{|c|}{F1} &S& Acc & P & R &\multicolumn{1}{|c|}{F1} &S& Acc& P & R &\multicolumn{1}{|c}{F1}\\
\hline
\multicolumn{3}{c|}{RNP$^*$} &10.0&-& 32.2 & 18.6 & 23.6  &10.0&-&44.8 &32.4& 37.6 & 10.0 & - & 24.6 & 23.5 & 24.0\\
\multicolumn{3}{c|}{INVRAT$^{*}$} & 10.0 & - & 42.6 &31.5 & 36.2 &10.0&- & 41.2 & 39.1 & 40.1&10.0&-&34.9&45.6&39.5\\
\multicolumn{3}{c|}{Inter-RAT$^{*}$} & 11.7 & -&66.0 &46.5 & 54.6 &11.7 & - & 55.4 & 47.5&51.1&12.6&-&34.6&48.2&40.2\\
\multicolumn{3}{c|}{MGR(ours)} &10.9&80.5&\textbf{87.5}&\textbf{51.7}&\textbf{65.0}&10.3&{89.7}&\textbf{78.7}&\textbf{52.2}&\textbf{62.8}&10.8&{86.0}&\textbf{65.6}&\textbf{57.1}&\textbf{61.1}
  \\\hline
 \multicolumn{3}{c|}{RNP$^*$} &20.0&-& 39.4 & 44.9 & 42.0  &20.0&-&37.5 &51.9& 43.5 & 20.0 & - & 21.6 & 38.9 & 27.8\\
\multicolumn{3}{c|}{INVRAT$^{*}$} & 20.0 & - & 58.9 &67.2 & 62.8 &20.0&- & 29.3 & 52.1 & 37.5&20.0&-&24.0&55.2&33.5\\
\multicolumn{3}{c|}{Inter-RAT$^{*}$} & 21.7 & -&62.0 &76.7 & 68.6 &20.4 & - & 44.2 & 65.4&52.8&20.8&-&26.3&59.1&36.4\\
\multicolumn{3}{c|}{MGR(ours)} &20.3&85.6&\textbf{76.3}&\textbf{83.6}&\textbf{79.8}&19.7&{89.6}&\textbf{64.4}&\textbf{81.3}&\textbf{71.9}&19.3&{89.3}&\textbf{47.1}&\textbf{73.1}&\textbf{57.3}
  \\\hline
 \multicolumn{3}{c|}{RNP$^*$} &30.0&-& 24.2 & 41.2 & 30.5  &30.0&-&27.1 &55.7& 36.4 & 30.0 & - & 15.4 & 42.2 & 22.6\\
\multicolumn{3}{c|}{INVRAT$^{*}$} & 30.0 & - & 41.5 &74.8 & 53.4 &30.0&- & 22.8 & 65.1 & 33.8&30.0&-&20.9&\textbf{71.6}&32.3\\
\multicolumn{3}{c|}{Inter-RAT$^{*}$} & 30.5 & -&48.1 &82.7 & 60.8 &29.4 & - & 37.9 & 72.0&49.6&30.4&-&21.8&{66.1}&{32.8}\\
\multicolumn{3}{c|}{MGR(ours)} &30.4&88.5&\textbf{57.2}&\textbf{93.9}&\textbf{71.1}&29.8&{91.6}&\textbf{45.8}&\textbf{87.4}&\textbf{60.1}&30.3&{89.3}&\textbf{27.3}&{66.5}&\textbf{38.7}
  \\\hline
\end{tabular}
}
    \caption{Results on \emph{correlated BeerAdvocate}. $``*"$: results obtained from Inter-RAT \citep{interventional}. }
    \label{tab:correlated beer}
\end{table*}

To support our claims, we conduct two practical experiments on \emph{decorrelated BeerAdvocate} dataset, where the main problem is degeneration. First, we compare the performance of MGR using one learning rate to MGR using separate learning rates. The results are shown in Figure~\ref{fig:nodis and onehead}(a). Although using separate learning rates does not help much in the relatively easy aspects including \emph{Appearance} and \emph{Aroma}, it makes a significant improvement in the hard \emph{Palate} aspect. Second, we compare the performance of keeping only one \emph{generator} for inference to averaging the results of multiple \emph{generators}, as shown in Figure~\ref{fig:nodis and onehead}(b). The results show that keeping only one \emph{generator} hardly influences the performance, which indicates that different \emph{generators} can finally converge to get the same outputs and only one \emph{generator} is required in inference time. We also show the differences in the rationales generated by different generators in Figure~\ref{fig:difference of rationales from different generators} of Appendix~\ref{app:rationale difference}.

\begin{table}[t]

    \centering
    \resizebox{0.99\columnwidth}{!}{
    \setlength\tabcolsep{3pt}
    \begin{tabular}{c|c c c c c}
    \hline
     & RNP & CAR& DMR & A2R & MGR(Ours)  \\
     \hline
  modules &1gen+1pred&1gen+2pred&1gen+3pred&1gen+2pred&3gen+1pred \\     parameters & 2$\times$ &3$\times$ &4$\times$ &3$\times$&4$\times$\\
  \hline
    \end{tabular}
    }
    \caption{The complexity
of different models. $``$gen$"$: generator. $``$pred$"$: predictor. }

    \label{tab:para}
\end{table}

\section{Experiments}
\subsection{Experimental Setup}
\textbf{Datasets}\label{sec:dataset}
1) \textbf{BeerAdvocate} \citep{beer} is a multi-aspect sentiment prediction dataset widely used in rationalization. There is a high correlation among the rating scores of different aspects in the same review, making the rationale selecting faces a severe spurious correlation. We use the original dataset to verify the effectiveness of MGR in dealing with spurious correlation and degeneration at the same time. In addition, following the previous work \citep{rnp,dmr,interlocking}, we use the subsets containing less spurious correlation \citep{rnp} to see the effectiveness in dealing with solitary degeneration.
2) \textbf{Hotel Reviews} \citep{hotel} is another multi-aspect sentiment classification dataset also widely used in rationalization. Each aspect itself can be seen as a dataset and is trained independently.

\begin{table*}[t]
\subtable[Normal experiments on \emph{decorrelated BeerAdvocate} ]{
    \centering

\resizebox{1.99\columnwidth}{!}{
    \begin{tabular}{ c c |c c| c c |c|c c| c c |c |c c| c c |c }
\hline
\multicolumn{2}{c|}{\multirow{2}{*}{Methods}} & \multicolumn{5}{c|}{Appearance} & \multicolumn{5}{c|}{Aroma} & \multicolumn{5}{c}{Palate}\\
\cline{3-17}
\multicolumn{2}{c|}{} &S& Acc & P & R &\multicolumn{1}{|c|}{F1} &S& Acc & P & R &\multicolumn{1}{|c|}{F1} &S& Acc& P & R &\multicolumn{1}{|c}{F1}\\
\hline
\multicolumn{2}{c|}{RNP$^*$} &os& 85.7 & 83.9 & 71.2 & 72.8 &os&84.2&73.6 &67.9& 65.9 & os & 83.8 & 55.5 & 54.3 & 51.0\\
\multicolumn{2}{c|}{re-DMR} &18.2&-& 71.1 &70.2 & 70.7 &15.4&-&59.8 & 58.9 & 59.3 &11.9&-&53.2 &50.9 & 52.0\\
\multicolumn{2}{c|}{re-A2R} & 18.4 & 83.9 & 72.7 &72.3 & 72.5 & 15.4 & 86.3 & 63.6 & 62.9&63.2&12.4&81.2&57.4&57.3&57.4\\
\multicolumn{2}{c|}{A2R$^*$} & os & 86.3 & 84.7 &71.2 & 72.9 & os & 84.9 & \textbf{79.3} & 71.3&70.0&os&84.0&64.2&60.9&58.0\\
\multicolumn{2}{c|}{MGR(ours)} &18.4&86.1&\textbf{83.9}&\textbf{83.5}&\textbf{83.7
}&15.6&86.6&76.6&\textbf{76.5}&\textbf{76.5}&12.4&85.1&\textbf{66.6}&\textbf{66.6}&\textbf{66.6}
  \\\hline
\end{tabular}
}
\label{tab:beer}
}

\subtable[Beer-Skewed in \emph{Palate} aspect of  \emph{decorrelated BeerAdvocate} ]{
    \centering

\resizebox{1.99\columnwidth}{!}{
\footnotesize
    \begin{tabular}{c |c| c c |c|c| c c |c|c| c c |c}
    \hline
     \multirow{2}{*}{Setting}& \multicolumn{4}{c|}{RNP$^*$} & \multicolumn{4}{c|}{A2R$^*$}& \multicolumn{4}{c}{MGR(ours)} \\
\cline{2-13}
{} &Acc&P & R &F1&Acc & P & R &F1&Acc&P & R &F1\\
\hline
\multicolumn{1}{c|}{skew10} &77.3&5.6 &7.4 & 5.5 &82.8&50.3&48.0&45.5 &82.0&\textbf{65.2}&\textbf{62.8}&\textbf{64.0}\\
\multicolumn{1}{c|}{skew15} &77.1&1.2 & 2.5 & 1.3 &80.9&30.2&29.9&27.7&77.4&\textbf{62.7}&\textbf{58.2}&\textbf{60.4}\\
\multicolumn{1}{c|}{skew20} &75.6&0.4 & 1.4 & 0.6 &76.7&0.4&1.6&0.6& 82.5 &\textbf{65.6}&\textbf{63.2}&\textbf{64.4}\\
\hline
    \end{tabular}
}
\label{tab:beer-skewed}
}

    \caption{The standard experiment and one synthetic experiment on \emph{decorrelated BeerAdvocate}. $``*"$: results from the paper of A2R. $``$re-$"$: our reimplemented methods. $``$os$"$: one sentence. }
    \label{tab:beer and beer-skewed}
\end{table*}

\begin{table*}[t]
    \centering

    \resizebox{1.99\columnwidth}{!}{
    \begin{tabular}{c c c |c c| c c |c|c c| c c |c |c c| c c |c }
\hline
\multicolumn{3}{c|}{\multirow{2}{*}{Methods}} & \multicolumn{5}{c|}{Location} & \multicolumn{5}{c|}{Service} & \multicolumn{5}{c}{Cleanliness}\\
\cline{4-18}
\multicolumn{3}{c|}{} &S& Acc & P & R &\multicolumn{1}{|c|}{F1} &S& Acc & P & R &\multicolumn{1}{|c|}{F1} &S& Acc& P & R &\multicolumn{1}{|c}{F1}\\
\hline
\multicolumn{3}{c|}{RNP$^*$} &10.9&-& 43.3 & 55.5 & 48.6  &11.0&-&40.0 &38.2& 39.1 & 10.6 & - & 30.5 & 36.0 & 33.0\\
\multicolumn{3}{c|}{CAR$^*$} &10.6&-& 46.6 &58.1 & 51.7 &11.7&-&40.7 & 41.4 & 41.1 &9.9&-&32.3&35.7 & 33.9\\
\multicolumn{3}{c|}{DMR$^{**}$} & 10.7 & - & 47.5 &60.1 & 53.1 &11.6&- & 43.0 & 43.6 & 43.3&10.3&-&31.4&36.4&33.7\\
\multicolumn{3}{c|}{re-A2R} &  8.5 &87.5 &43.1 &43.2 & 43.1 &11.4 & 96.5 & 37.3 & 37.2&37.2&8.9&94.5&33.2&33.3&33.3\\
\multicolumn{3}{c|}{MGR(ours)} &9.7&97.5&\textbf{52.5}&\textbf{60.5}&\textbf{56.2}&11.8&{96.5}&\textbf{45.0}&\textbf{46.4}&\textbf{45.7}&10.5&{96.5}&\textbf{37.6}&\textbf{44.5}&\textbf{40.7}
  \\\hline
\end{tabular}
}
    \caption{Results on \emph{HotelReview}. Each aspect is trained independently.  $``*"$: results from the paper of CAR \cite{car}, $``**"$: results from the paper of DMR. $``$re-$"$: our reimplemented method. }
    \label{tab:hotel}
\end{table*}

\noindent\textbf{Baselines and implementation details}.\label{sec:details}
In parcitce, we set $n=3$ (the number of generators) as for our MGR as a performance-time trade-off.
We compare MGR to the vanilla RNP \citep{rnp} and several latest published methods that achieve state-of-the-art results: INVRAT \citep{invarant}, DMR \citep{dmr}, A2R \citep{interlocking}, Inter-RAT \citep{interventional}, all of which have been specified in Section~\ref{sec:related work}. 
Following the commonly used rationalization settings \cite{car,rethinking,invarant,dmr,interlocking,interventional}, we use the 100-dimension Glove \citep{glove} as the word embedding and 200-dimension GRUs to get the text representation. We do not use BERT \citep{bert} because it is still a challenging task to finetune large pretrained models on the RNP cooperative framework (see Table 4 in \citet{danqi} and Appendix~\ref{app:bert}).
We use Adam \citep{adam} as the optimizer. 
All the baselines are tuned multiple times manually to find the best hyperparameters. The complexity of different models are shown in  Table~\ref{tab:para}
All of the models are trained on a RTX3090 GPU. More details are in Appendix~\ref{app:implemenration details}.

\noindent\textbf{Metrics}.
All the methods get similar predictive accuracy. Following \citep{invarant,dmr,interlocking,interventional}, we mainly focus on the quality of rationales, which is measured by the overlap between the model-selected tokens and human-annotated tokens. $P,R,F1$ indicate the precision, recall, and F1 score, respectively. $S$ indicates the average sparsity of the selected rationales, i.e., the percentage of selected tokens to the whole texts. $Acc$ indicates the predictive accuracy of the test set.

\subsection{Results}
We first conduct an experiment on the \emph{correlated BeerAdvocate} dataset, where the problems of degeneration and spurious correlation both may damage the rationale quality. Methods that achieve the state-of-the-art results on this dataset are INVRAT \citep{invarant} and Inter-RAT \citep{interventional}. We tune $s$ in Equation~\ref{eqa:regular} to get similar rationale sparsity as previous methods do. The results are shown in Table~\ref{tab:correlated beer}. We improve the F1 score by up to $20.9\%$ (\emph{Palate} aspect with $20\%$ sparsity) over the latest SOTA. Besides, except the \emph{Palate} aspect with $30\%$ sparsity, we get over $10\%$ improvements under all the other settings.

We then conduct an experiment on the \emph{decorrelated BeerAdvocate} dataset, where the main problem is degeneration. Methods that achieve the state-of-the-art results on this dataset are DMR \citep{dmr} and A2R \citep{interlocking}. Since the rationales of \emph{BeerAdvocate} on a sentence level, A2R in its original paper does sentence level selection (i.e., selecting one sentence as the rationale) on this dataset. We also reimplement A2R according to its source codes to do the token-level selection. The results are shown in Table~\ref{tab:beer}. The sparsity is set to be close to that of the human-annotated rationales. We beat all the methods in terms of F1 score. We do not get as significant improvements as those in Table~\ref{tab:correlated beer} because the spurious correlation is removed manually in this dataset. But we still get up to $10.8\%$ (\emph{Appearance} aspect) improvements as compared to the SOTA.

To show the generalizability of our method, we further conduct an experiment on \emph{HotelReviews}. Methods that achieve the state-of-the-art results on this dataset are DMR \citep{dmr} and CAR \citep{car}. We also beat all the baselines and get up to $6.8\%$ (\emph{Cleanliness} aspect) improvements on this dataset.

\begin{table}[t]
    \centering
    \resizebox{0.99\columnwidth}{!}{
    \begin{tabular}{c c c |c c| c c |c }
\hline
\multicolumn{3}{c|}{\multirow{2}{*}{Methods}} & \multicolumn{5}{c}{Appearance} \\
\cline{4-8}
\multicolumn{3}{c|}{} &S& Acc & P & R &\multicolumn{1}{|c}{F1} \\
\hline

\multicolumn{3}{c|}{MGR(n=5)} &19.2&86.3&{83.8}&{86.8}&{85.3}\\
\multicolumn{3}{c|}{MGR(n=7)} &19.6&87.0&{83.5}&{88.3}&{85.8}\\
\multicolumn{3}{c|}{MGR(n=9)} &19.4&86.0&{83.6}&{87.7}&{85.6}

  \\\hline
\end{tabular}
}
    \caption{Results of MGR with different numbers of generators. The dataset is the \emph{Appearance} aspect of \emph{correlated BeerAdvocate}. The sparsity is $S\approx 20$. }
    \label{tab: beer with different n}
\end{table}

\noindent\textbf{Results of MGR with different numbers of generators}. 
Although we set $n=3$ in our previous experiments, we also show the results of our MGR with different values of $n$ in Table~\ref{tab: beer with different n}. When $n$ grows, the results are somewhat better than those of $n=3$. However, $n=3$ yields the most improvements per additional generator and proved to be a good performance-cost trade-off. And note that, having too many generators may not always result in better outcomes, because the learning rate for the $i$-th generator, which is $i \times \eta$, may become too large for stable training.

\noindent\textbf{Beer-Skewed}. To show that our MGR does not suffer from the degeneration problem, we conduct the same synthetic experiment that deliberately induces degeneration as \citet{interlocking} did. The details of the experimental setup are in Appendix~\ref{app:skewpred}. We use the relatively harder \emph{Palate} aspect \cite{interlocking}. The results are shown in Table~\ref{tab:beer-skewed}. The results of RNP and A2R are obtained from \citep{interlocking}.  For all the settings, we outperform both RNP and A2R. Especially, for \emph{skew}$20$, RNP and A2R can not work at all while our MGR is only slightly influenced as compared to the corresponding result in Table~\ref{tab:beer}.

\begin{figure}
    \centering
    \includegraphics[width=0.6\columnwidth]{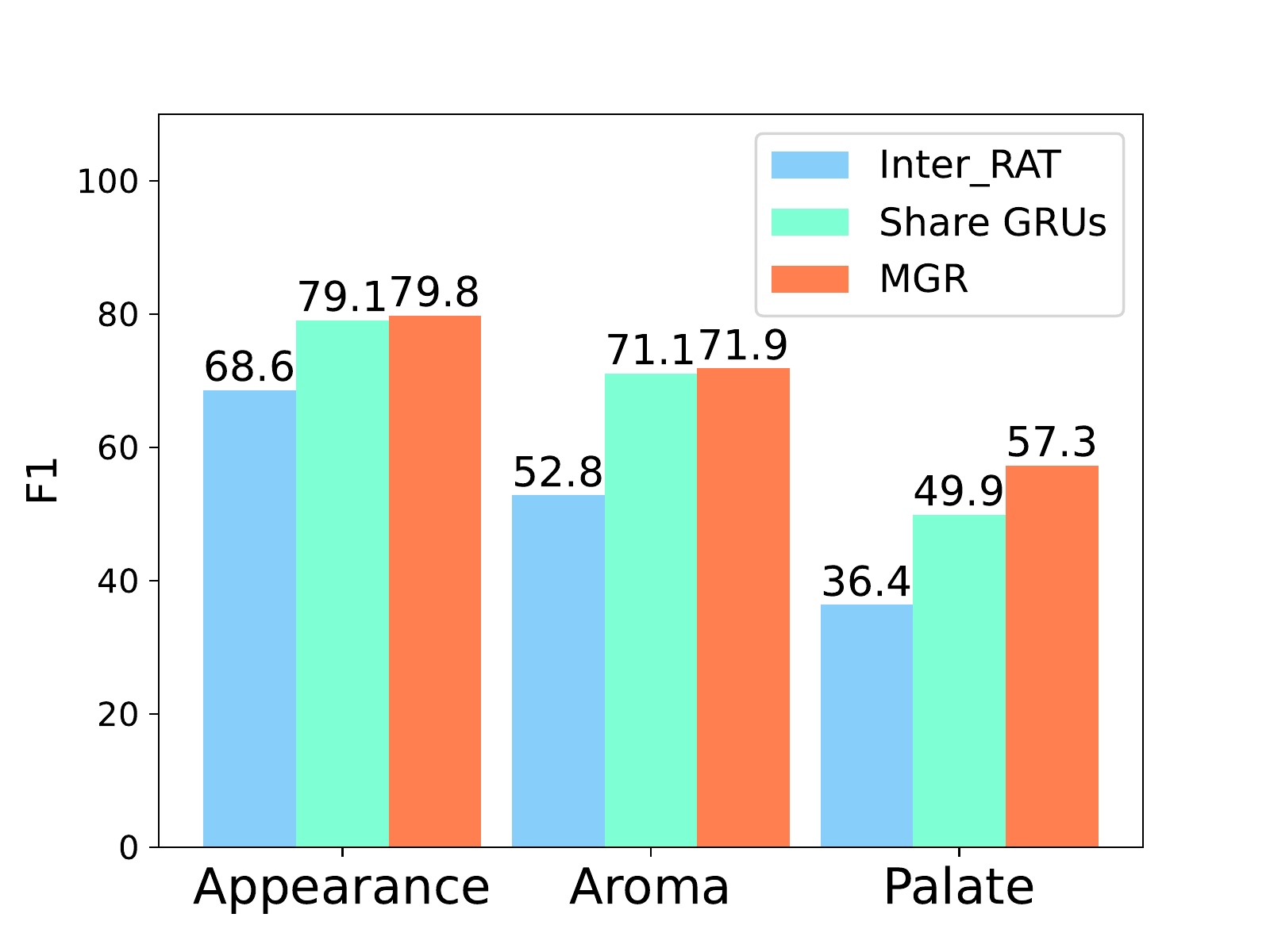}
    \caption{The comparison of sharing the encoders (GRUs) of MGR's generator.
 }
    \label{fig:share gru sp20}
\end{figure}

\noindent\textbf{Sharing encoders between generators}. The major limitation of MGR is the increased computational costs. One plausible trade-off method may be sharing some parameters between the generators. We conduct an experiment where we share the generators' GRUs but keep their own linear heads. Figure~\ref{fig:share gru sp20} shows the results on \emph{correlated BeerAdvocate} with sparsity around $20\%$. More results are in Appendix~\ref{app:share grus}. Although simply sharing the generators' encoders sometimes cause damage to the performance of MGR, it still outperforms the state-of-the-art method Inter\_RAT. We leave how to better decrease the computational costs without hurting the model performance as future work.

\section{Conclusion and future work}
In this paper, we design a new framework MGR to simultaneously tackle the two major challenges including the spurious correlation and degeneration in the self-explaining rationalization schemes. Specifically, we propose leveraging multiple generators to select rationales such that the predictor can have access to more meaningful rationales stably. We theoretically show that the proposed method can solve the two problems. Finally, empirical results conducted on various datasets demonstrate the great effectiveness of our proposed method.

\section*{Limitations}
More generators bring significant benefits to the model performance to our MGR, but the training cost is also increased with the number of generators growing. Although we have verified that we only need to keep one generator during test, there is no denying that the training cost is still an important problem. In the future, we will explore some methods like multi-task learning and model fusion, to reduce the model complexity.   

\section*{Acknowledgements}
This work is supported by National Natural Science Foundation of China under grants U1836204, U1936108, 62206102,  and Science and Technology Support Program of Hubei Province under grant 2022BAA046. We thank the anonymous reviewers for their valuable comments on improving the quality of this paper.

\bibliography{anthology,custom}

\begin{thebibliography}{31}
\expandafter\ifx\csname natexlab\endcsname\relax\def\natexlab#1{#1}\fi

\bibitem[{Ainsworth et~al.(2022)Ainsworth, Hayase, and Srinivasa}]{gitrebasin}
Samuel~K. Ainsworth, Jonathan Hayase, and Siddhartha~S. Srinivasa. 2022.
\newblock \href {https://doi.org/10.48550/arXiv.2209.04836} {Git re-basin:
  Merging models modulo permutation symmetries}.
\newblock \emph{CoRR}, abs/2209.04836.

\bibitem[{Bao et~al.(2018)Bao, Chang, Yu, and Barzilay}]{2018rationalegumble}
Yujia Bao, Shiyu Chang, Mo~Yu, and Regina Barzilay. 2018.
\newblock \href {https://doi.org/10.18653/v1/d18-1216} {Deriving machine
  attention from human rationales}.
\newblock In \emph{Proceedings of the 2018 Conference on Empirical Methods in
  Natural Language Processing, Brussels, Belgium, October 31 - November 4,
  2018}, pages 1903--1913. Association for Computational Linguistics.

\bibitem[{Bastings et~al.(2019)Bastings, Aziz, and Titov}]{hardkuma}
Jasmijn Bastings, Wilker Aziz, and Ivan Titov. 2019.
\newblock \href {https://doi.org/10.18653/v1/p19-1284} {Interpretable neural
  predictions with differentiable binary variables}.
\newblock In \emph{Proceedings of the 57th Conference of the Association for
  Computational Linguistics, {ACL} 2019, Florence, Italy, July 28- August 2,
  2019, Volume 1: Long Papers}, pages 2963--2977. Association for Computational
  Linguistics.

\bibitem[{Breiman(1996)}]{bagging}
Leo Breiman. 1996.
\newblock \href {https://doi.org/10.1007/BF00058655} {Bagging predictors}.
\newblock \emph{Mach. Learn.}, 24(2):123--140.

\bibitem[{Chang et~al.(2019)Chang, Zhang, Yu, and Jaakkola}]{car}
Shiyu Chang, Yang Zhang, Mo~Yu, and Tommi~S. Jaakkola. 2019.
\newblock \href
  {https://proceedings.neurips.cc/paper/2019/hash/5ad742cd15633b26fdce1b80f7b39f7c-Abstract.html}
  {A game theoretic approach to class-wise selective rationalization}.
\newblock In \emph{Advances in Neural Information Processing Systems 32: Annual
  Conference on Neural Information Processing Systems 2019, NeurIPS 2019,
  December 8-14, 2019, Vancouver, BC, Canada}, pages 10055--10065.

\bibitem[{Chang et~al.(2020)Chang, Zhang, Yu, and Jaakkola}]{invarant}
Shiyu Chang, Yang Zhang, Mo~Yu, and Tommi~S. Jaakkola. 2020.
\newblock \href {http://proceedings.mlr.press/v119/chang20c.html} {Invariant
  rationalization}.
\newblock In \emph{Proceedings of the 37th International Conference on Machine
  Learning, {ICML} 2020, 13-18 July 2020, Virtual Event}, volume 119 of
  \emph{Proceedings of Machine Learning Research}, pages 1448--1458. {PMLR}.

\bibitem[{Chen et~al.(2022)Chen, He, Narasimhan, and Chen}]{danqi}
Howard Chen, Jacqueline He, Karthik Narasimhan, and Danqi Chen. 2022.
\newblock \href {https://doi.org/10.18653/v1/2022.naacl-main.278} {Can
  rationalization improve robustness?}
\newblock In \emph{Proceedings of the 2022 Conference of the North American
  Chapter of the Association for Computational Linguistics: Human Language
  Technologies, {NAACL} 2022, Seattle, WA, United States, July 10-15, 2022},
  pages 3792--3805. Association for Computational Linguistics.

\bibitem[{Cho et~al.(2014)Cho, van Merrienboer, G{\"{u}}l{\c{c}}ehre, Bahdanau,
  Bougares, Schwenk, and Bengio}]{gru}
Kyunghyun Cho, Bart van Merrienboer, {\c{C}}aglar G{\"{u}}l{\c{c}}ehre, Dzmitry
  Bahdanau, Fethi Bougares, Holger Schwenk, and Yoshua Bengio. 2014.
\newblock \href {https://doi.org/10.3115/v1/d14-1179} {Learning phrase
  representations using {RNN} encoder-decoder for statistical machine
  translation}.
\newblock In \emph{Proceedings of the 2014 Conference on Empirical Methods in
  Natural Language Processing, {EMNLP} 2014, October 25-29, 2014, Doha, Qatar,
  {A} meeting of SIGDAT, a Special Interest Group of the {ACL}}, pages
  1724--1734. {ACL}.

\bibitem[{Deng et~al.(2023)Deng, Li, Guo, and Li}]{deng2023multi}
Zhiying Deng, Jianjun Li, Zhiqiang Guo, and Guohui Li. 2023.
\newblock Multi-aspect interest neighbor-augmented network for next-basket
  recommendation.
\newblock In \emph{ICASSP 2023-2023 IEEE International Conference on Acoustics,
  Speech and Signal Processing (ICASSP)}, pages 1--5. IEEE.

\bibitem[{Devlin et~al.(2019)Devlin, Chang, Lee, and Toutanova}]{bert}
Jacob Devlin, Ming{-}Wei Chang, Kenton Lee, and Kristina Toutanova. 2019.
\newblock \href {https://doi.org/10.18653/v1/n19-1423} {{BERT:} pre-training of
  deep bidirectional transformers for language understanding}.
\newblock In \emph{Proceedings of the 2019 Conference of the North American
  Chapter of the Association for Computational Linguistics: Human Language
  Technologies, {NAACL-HLT} 2019, Minneapolis, MN, USA, June 2-7, 2019, Volume
  1 (Long and Short Papers)}, pages 4171--4186. Association for Computational
  Linguistics.

\bibitem[{Guerreiro and Martins(2021)}]{spec}
Nuno~Miguel Guerreiro and Andr{\'{e}} F.~T. Martins. 2021.
\newblock \href {https://doi.org/10.18653/v1/2021.emnlp-main.525} {{SPECTRA:}
  sparse structured text rationalization}.
\newblock In \emph{Proceedings of the 2021 Conference on Empirical Methods in
  Natural Language Processing, {EMNLP} 2021, Virtual Event / Punta Cana,
  Dominican Republic, 7-11 November, 2021}, pages 6534--6550. Association for
  Computational Linguistics.

\bibitem[{Hase et~al.(2020)Hase, Zhang, Xie, and Bansal}]{leakage}
Peter Hase, Shiyue Zhang, Harry Xie, and Mohit Bansal. 2020.
\newblock Leakage-adjusted simulatability: Can models generate non-trivial
  explanations of their behavior in natural language?
\newblock In \emph{Findings of the Association for Computational Linguistics:
  EMNLP 2020}, pages 4351--4367.

\bibitem[{Huang et~al.(2021)Huang, Chen, Du, and Yang}]{dmr}
Yongfeng Huang, Yujun Chen, Yulun Du, and Zhilin Yang. 2021.
\newblock \href {https://ojs.aaai.org/index.php/AAAI/article/view/17547}
  {Distribution matching for rationalization}.
\newblock In \emph{Thirty-Fifth {AAAI} Conference on Artificial Intelligence,
  {AAAI} 2021, Thirty-Third Conference on Innovative Applications of Artificial
  Intelligence, {IAAI} 2021, The Eleventh Symposium on Educational Advances in
  Artificial Intelligence, {EAAI} 2021, Virtual Event, February 2-9, 2021},
  pages 13090--13097. {AAAI} Press.

\bibitem[{Jain et~al.(2020)Jain, Wiegreffe, Pinter, and
  Wallace}]{jain2020faith}
Sarthak Jain, Sarah Wiegreffe, Yuval Pinter, and Byron~C. Wallace. 2020.
\newblock \href {https://doi.org/10.18653/v1/2020.acl-main.409} {Learning to
  faithfully rationalize by construction}.
\newblock In \emph{Proceedings of the 58th Annual Meeting of the Association
  for Computational Linguistics, {ACL} 2020, Online, July 5-10, 2020}, pages
  4459--4473. Association for Computational Linguistics.

\bibitem[{Kingma and Ba(2015)}]{adam}
Diederik~P. Kingma and Jimmy Ba. 2015.
\newblock \href {http://arxiv.org/abs/1412.6980} {Adam: {A} method for
  stochastic optimization}.
\newblock In \emph{3rd International Conference on Learning Representations,
  {ICLR} 2015, San Diego, CA, USA, May 7-9, 2015, Conference Track
  Proceedings}.

\bibitem[{Lei et~al.(2016)Lei, Barzilay, and Jaakkola}]{rnp}
Tao Lei, Regina Barzilay, and Tommi~S. Jaakkola. 2016.
\newblock \href {https://doi.org/10.18653/v1/d16-1011} {Rationalizing neural
  predictions}.
\newblock In \emph{Proceedings of the 2016 Conference on Empirical Methods in
  Natural Language Processing, {EMNLP} 2016, Austin, Texas, USA, November 1-4,
  2016}, pages 107--117. The Association for Computational Linguistics.

\bibitem[{Liu et~al.(2022)Liu, Wang, Wang, Li, Yue, and Zhang}]{liufr}
Wei Liu, Haozhao Wang, Jun Wang, Ruixuan Li, Chao Yue, and YuanKai Zhang. 2022.
\newblock \href
  {https://proceedings.neurips.cc/paper_files/paper/2022/file/2e0bd92a1d3600d4288df51ac5e6be5f-Paper-Conference.pdf}
  {Fr: Folded rationalization with a unified encoder}.
\newblock In \emph{Advances in Neural Information Processing Systems},
  volume~35. Curran Associates, Inc.

\bibitem[{Liu et~al.(2023)Liu, Wang, Wang, Li, Qiu, Zhang, Han, and
  Zou}]{liudr}
Wei Liu, Jun Wang, Haozhao Wang, Ruixuan Li, Yang Qiu, Yuankai Zhang, Jie Han,
  and Yixiong Zou. 2023.
\newblock \href {https://doi.org/10.48550/arXiv.2305.13599} {Decoupled
  rationalization with asymmetric learning rates: {A} flexible lipschitz
  restraint}.
\newblock \emph{CoRR}, abs/2305.13599.

\bibitem[{McAuley et~al.(2012)McAuley, Leskovec, and Jurafsky}]{beer}
Julian~J. McAuley, Jure Leskovec, and Dan Jurafsky. 2012.
\newblock \href {https://doi.org/10.1109/ICDM.2012.110} {Learning attitudes and
  attributes from multi-aspect reviews}.
\newblock In \emph{12th {IEEE} International Conference on Data Mining, {ICDM}
  2012, Brussels, Belgium, December 10-13, 2012}, pages 1020--1025. {IEEE}
  Computer Society.

\bibitem[{Paranjape et~al.(2020)Paranjape, Joshi, Thickstun, Hajishirzi, and
  Zettlemoyer}]{informationbottle}
Bhargavi Paranjape, Mandar Joshi, John Thickstun, Hannaneh Hajishirzi, and Luke
  Zettlemoyer. 2020.
\newblock \href {https://doi.org/10.18653/v1/2020.emnlp-main.153} {An
  information bottleneck approach for controlling conciseness in rationale
  extraction}.
\newblock In \emph{Proceedings of the 2020 Conference on Empirical Methods in
  Natural Language Processing, {EMNLP} 2020, Online, November 16-20, 2020},
  pages 1938--1952. Association for Computational Linguistics.

\bibitem[{Pennington et~al.(2014)Pennington, Socher, and Manning}]{glove}
Jeffrey Pennington, Richard Socher, and Christopher~D. Manning. 2014.
\newblock \href {https://doi.org/10.3115/v1/d14-1162} {Glove: Global vectors
  for word representation}.
\newblock In \emph{Proceedings of the 2014 Conference on Empirical Methods in
  Natural Language Processing, {EMNLP} 2014, October 25-29, 2014, Doha, Qatar,
  {A} meeting of SIGDAT, a Special Interest Group of the {ACL}}, pages
  1532--1543. {ACL}.

\bibitem[{Plyler et~al.(2021)Plyler, Green, and Chi}]{counter}
Mitchell Plyler, Michael Green, and Min Chi. 2021.
\newblock \href
  {https://proceedings.neurips.cc/paper/2021/hash/f0f800c92d191d736c4411f3b3f8ef4a-Abstract.html}
  {Making a (counterfactual) difference one rationale at a time}.
\newblock In \emph{Advances in Neural Information Processing Systems 34: Annual
  Conference on Neural Information Processing Systems 2021, NeurIPS 2021,
  December 6-14, 2021, virtual}, pages 28701--28713.

\bibitem[{Rajagopal et~al.(2021)Rajagopal, Balachandran, Hovy, and
  Tsvetkov}]{concept}
Dheeraj Rajagopal, Vidhisha Balachandran, Eduard~H Hovy, and Yulia Tsvetkov.
  2021.
\newblock \href {https://doi.org/10.18653/v1/2021.emnlp-main.64}
  {{SELFEXPLAIN}: A self-explaining architecture for neural text classifiers}.
\newblock In \emph{Proceedings of the 2021 Conference on Empirical Methods in
  Natural Language Processing}, pages 836--850, Online and Punta Cana,
  Dominican Republic. Association for Computational Linguistics.

\bibitem[{Schapire(1999)}]{boosting}
Robert~E. Schapire. 1999.
\newblock \href {http://ijcai.org/Proceedings/99-2/Papers/103.pdf} {A brief
  introduction to boosting}.
\newblock In \emph{Proceedings of the Sixteenth International Joint Conference
  on Artificial Intelligence, {IJCAI} 99, Stockholm, Sweden, July 31 - August
  6, 1999. 2 Volumes, 1450 pages}, pages 1401--1406. Morgan Kaufmann.

\bibitem[{Singh and Jaggi(2020)}]{fusion}
Sidak~Pal Singh and Martin Jaggi. 2020.
\newblock \href
  {https://proceedings.neurips.cc/paper/2020/hash/fb2697869f56484404c8ceee2985b01d-Abstract.html}
  {Model fusion via optimal transport}.
\newblock In \emph{Advances in Neural Information Processing Systems 33: Annual
  Conference on Neural Information Processing Systems 2020, NeurIPS 2020,
  December 6-12, 2020, virtual}.

\bibitem[{Wang et~al.(2010)Wang, Lu, and Zhai}]{hotel}
Hongning Wang, Yue Lu, and Chengxiang Zhai. 2010.
\newblock \href {https://doi.org/10.1145/1835804.1835903} {Latent aspect rating
  analysis on review text data: a rating regression approach}.
\newblock In \emph{Proceedings of the 16th {ACM} {SIGKDD} International
  Conference on Knowledge Discovery and Data Mining, Washington, DC, USA, July
  25-28, 2010}, pages 783--792. {ACM}.

\bibitem[{Wolpert(1992)}]{stacked}
David~H. Wolpert. 1992.
\newblock \href {https://doi.org/10.1016/S0893-6080(05)80023-1} {Stacked
  generalization}.
\newblock \emph{Neural Networks}, 5(2):241--259.

\bibitem[{Yu et~al.(2019)Yu, Chang, Zhang, and Jaakkola}]{rethinking}
Mo~Yu, Shiyu Chang, Yang Zhang, and Tommi~S. Jaakkola. 2019.
\newblock \href {https://doi.org/10.18653/v1/D19-1420} {Rethinking cooperative
  rationalization: Introspective extraction and complement control}.
\newblock In \emph{Proceedings of the 2019 Conference on Empirical Methods in
  Natural Language Processing and the 9th International Joint Conference on
  Natural Language Processing, {EMNLP-IJCNLP} 2019, Hong Kong, China, November
  3-7, 2019}, pages 4092--4101. Association for Computational Linguistics.

\bibitem[{Yu et~al.(2021)Yu, Zhang, Chang, and Jaakkola}]{interlocking}
Mo~Yu, Yang Zhang, Shiyu Chang, and Tommi~S. Jaakkola. 2021.
\newblock \href
  {https://proceedings.neurips.cc/paper/2021/hash/6a711a119a8a7a9f877b5f379bfe9ea2-Abstract.html}
  {Understanding interlocking dynamics of cooperative rationalization}.
\newblock In \emph{Advances in Neural Information Processing Systems 34: Annual
  Conference on Neural Information Processing Systems 2021, NeurIPS 2021,
  December 6-14, 2021, virtual}, pages 12822--12835.

\bibitem[{Yuan et~al.(2022)Yuan, Cai, Hu, Wang, and Ji}]{GDM}
Hao Yuan, Lei Cai, Xia Hu, Jie Wang, and Shuiwang Ji. 2022.
\newblock \href {https://doi.org/10.1109/TPAMI.2020.3028783} {Interpreting
  image classifiers by generating discrete masks}.
\newblock \emph{IEEE Transactions on Pattern Analysis and Machine
  Intelligence}, 44(4):2019--2030.

\bibitem[{Yue et~al.(2023)Yue, Liu, Wang, An, Du, and Huang}]{interventional}
Linan Yue, Qi~Liu, Li~Wang, Yanqing An, Yichao Du, and Zhenya Huang. 2023.
\newblock \href {https://openreview.net/forum?id=KoEa6h1o6D1} {Interventional
  rationalization}.

\end{thebibliography}
\bibliographystyle{acl_natbib}

\clearpage
\appendix

\section{More Results}
\subsection{More implementation details}\label{app:implemenration details}
To the best of our knowledge, both datasets are sufficiently anonymized to make identification of individuals impossible without significant effort. Both datasets are in English. For \emph{correlated BeerAdvocate}, we preprocess the data in the same way as \citet{interventional}. For \emph{decorrelated BeerAdvocate} and \emph{Hotel Reviews}, we preprocess them in the same way as \citet{dmr}. The maximum text length is set to 256. More statistics of the datasets are in Table~\ref{tab:dataset}.

\begin{table}[t]
   \centering
  \resizebox{0.99\columnwidth}{!}{  
    \begin{tabular}{c l| c c| c c| c c c}
    \hline
         \multicolumn{2}{c|}{\multirow{2}{*}{Datasets}}&\multicolumn{2}{c|}{Train}&\multicolumn{2}{c|}{Dev}&\multicolumn{3}{c}{Annotation}  \\
         \multicolumn{2}{c|}{}& Pos&Neg&Pos&Neg&Pos&Neg&Sparsity\\
         \hline\hline
        \multirow{3}{*}{Beer}&Appearance&202385&12897 &28488&1318&923&13&18.5\\
        {}&Aroma&172299&30564&24494&3396&848&29&15.6\\
        {}&Palate&176038&27639&24837&3203&785&20&12.4\\
        \hline\hline
        \multirow{3}{*}{Beer*}&Appearance&16891&16891 &6628&2103&923&13&18.5\\
        {}&Aroma&15169&15169&6579&2218&848&29&15.6\\
        {}&Palate&13652&13652&6740&2000&785&20&12.4\\
        \hline\hline
        \multirow{3}{*}{Hotel}&Location&7236&7236 &906&906&104&96&8.5\\
        {}&Service&50742&50742&6344&6344&101&99&11.5\\
        {}&Cleanliness&75049&75049&9382&9382&99&101&8.9\\
        \hline
    \end{tabular}
    }
    \caption{Statistics of datasets used in this paper. *: the decorrelated BeerAdvocate.}
    \label{tab:dataset}
\end{table}

Some previous methods needs very careful hyper-parameter tuning. To make fair comparisons, most results of the baselines are copied from previous papers. But some settings are not unified, so we also reimplement them according to their source codes.

For DMR, we adopt its source code and adjust its sparsity constraint to get a sparsity similar to the annotated rationales. For A2R, we re-implement it to do token-level selection as other methods do. 

The hyper-parameters of reimplemented models are manually tuned multiple times to get the best results. For our MGR, the early stopping technique is conducted according to the predictive accuracy of the development set. For our reimplemented DMR and A2R, although we have tried our best to tune the hyper-parameters, chances are that the hyper-parameters are not the best. To compensate for this potential issue, we do the test after every training epoch and choose their best results when they get the best $F1$ score on the test set.

The random seed is kept the same across all the experiments rather than manually selected. We think the experiments with one same random seed on multiple different settings and different datasets are enough to show the stability of our method. We also provide the standard deviations with running the experiments in Table~\ref{tab:correlated beer} with five different random seeds. The standard deviations are shown in Table~\ref{tab:correlated beer seed}.

\begin{table*}[t]
    \centering
    \resizebox{1.99\columnwidth}{!}{
    \begin{tabular}{l l l |l l| l l |l|l l| l l |l |l l| l l |l }
\hline
\multicolumn{3}{c|}{\multirow{2}{*}{Methods}} & \multicolumn{5}{c|}{Appearance} & \multicolumn{5}{c|}{Aroma} & \multicolumn{5}{c}{Palate}\\
\cline{4-18}
\multicolumn{3}{c|}{} &S& Acc & P & R &\multicolumn{1}{|l|}{F1} &S& Acc & P & R &\multicolumn{1}{|l|}{F1} &S& Acc& P & R &\multicolumn{1}{|l}{F1}\\
\hline
\multicolumn{3}{c|}{MGR(Table~\ref{tab:correlated beer})} &10.9&80.5& {87.5}& {51.7}& {65.0}&10.3&{89.7}& {78.7}& {52.2}& {62.8}&10.8&{86.0}& {65.6}& {57.1}& {61.1}\\
\multicolumn{3}{c|}{MGR$_{{\pm{std}}}$} &11.0$_{\pm{0.1}}$&80.1$_{\pm{0.7}}$& {85.6$_{\pm{1.4}}$}& {50.9$_{\pm{0.9}}$}& {63.8$_{\pm{1.0}}$}&9.7$_{\pm{0.5}}$&{88.2$_{\pm{1.7}}$}& {80.6$_{\pm{2.7}}$}& {50.3$_{\pm{1.6}}$}& {61.9$_{\pm{1.1}}$}&10.6$_{\pm{0.2}}$&{84.9$_{\pm{1.0}}$}& {62.8$_{\pm{2.2}}$}& {53.5$_{\pm{2.4}}$}& {57.9$_{\pm{2.3}}$}
  \\\hline

\multicolumn{3}{c|}{MGR(Table~\ref{tab:correlated beer})} &20.3&85.6& {76.3}& {83.6}& {79.8}&19.7&{89.6}& {64.4}& {81.3}& {71.9}&19.3&{89.3}& {47.1}& {73.1}& {57.3}\\
\multicolumn{3}{c|}{MGR$_{\pm{std}}$} &19.8$_{\pm{0.3}}$&86.7$_{\pm{1.1}}$& {79.4$_{\pm{1.9}}$}& {84.9$_{\pm{1.2}}$}& {82.1$_{\pm{1.5}}$}&19.3$_{\pm{0.3}}$&{88.6$_{\pm{0.8}}$}& {65.8$_{\pm{0.9}}$}& {81.3$_{\pm{0.8}}$}& {72.7$_{\pm{0.6}}$}&19.6$_{\pm{0.7}}$&{88.4$_{\pm{1.1}}$}& {46.3$_{\pm{1.9}}$}& {72.8$_{\pm{1.6}}$}& {56.6$_{\pm{1.8}}$}
  \\\hline
\multicolumn{3}{c|}{MGR(Table~\ref{tab:correlated beer})} &30.4&88.5& {57.2}& {93.9}& {71.1}&29.8&{91.6}& {45.8}& {87.4}& {60.1}&30.3&{89.3}& {27.3}&{66.5}& {38.7}\\
\multicolumn{3}{c|}{MGR$_{\pm{std}}$} &29.4$_{\pm{0.6}}$&87.0$_{\pm{1.5}}$& {57.8$_{\pm{0.4}}$}& {91.5$_{\pm{1.4}}$}& {70.8$_{\pm{0.3}}$}&29.6$_{\pm{0.4}}$&{89.5$_{\pm{1.5}}$}& {46.5$_{\pm{1.0}}$}& {88.8$_{\pm{1.8}}$}& {61.0$_{\pm{1.1}}$}&29.9$_{\pm{0.9}}$&{88.3$_{\pm{1.6}}$}& {26.4$_{\pm{1.1}}$}&{63.5$_{\pm{2.2}}$}& {37.3$_{\pm{1.4}}$}
  \\\hline
\end{tabular}
}
    \caption{The standard deviations of MGR on \emph{correlated BeerAdvocate} with five different random seeds. }
    \label{tab:correlated beer seed}
\end{table*}

\subsection{Discussion on BERT encoder}\label{app:bert}
 In the field of rationalization, researchers generally focus on frameworks of the models and the methodology. Methods most related to our work do not use Bert or other pre-trained encoders \cite{car,invarant,dmr,rethinking,interlocking,interventional}. We use GRUs and GloVe to ensure the same experimental setup as our baselines for a fair comparison.

   \begin{table}[t]
   \centering
   \setlength\tabcolsep{2pt}
    \begin{tabular}{c |c| c}
    \hline
    Methods & Beer-Appearance &Hotel-Cleanliness\\
    \hline
    VIB&20.5&23.5\\
    SPECTRA&28.6&19.5\\
    \hline
         
    \end{tabular}
    \caption{Results with BERT. VIB: \citet{informationbottle}, SPECTRA: \citet{spec}. The results are from Table 4 of \citet{danqi}. The metric is F1 score.}
    \label{tab:res_bert}
\end{table}

\begin{figure}
    \centering

    \includegraphics[width=0.7\columnwidth]{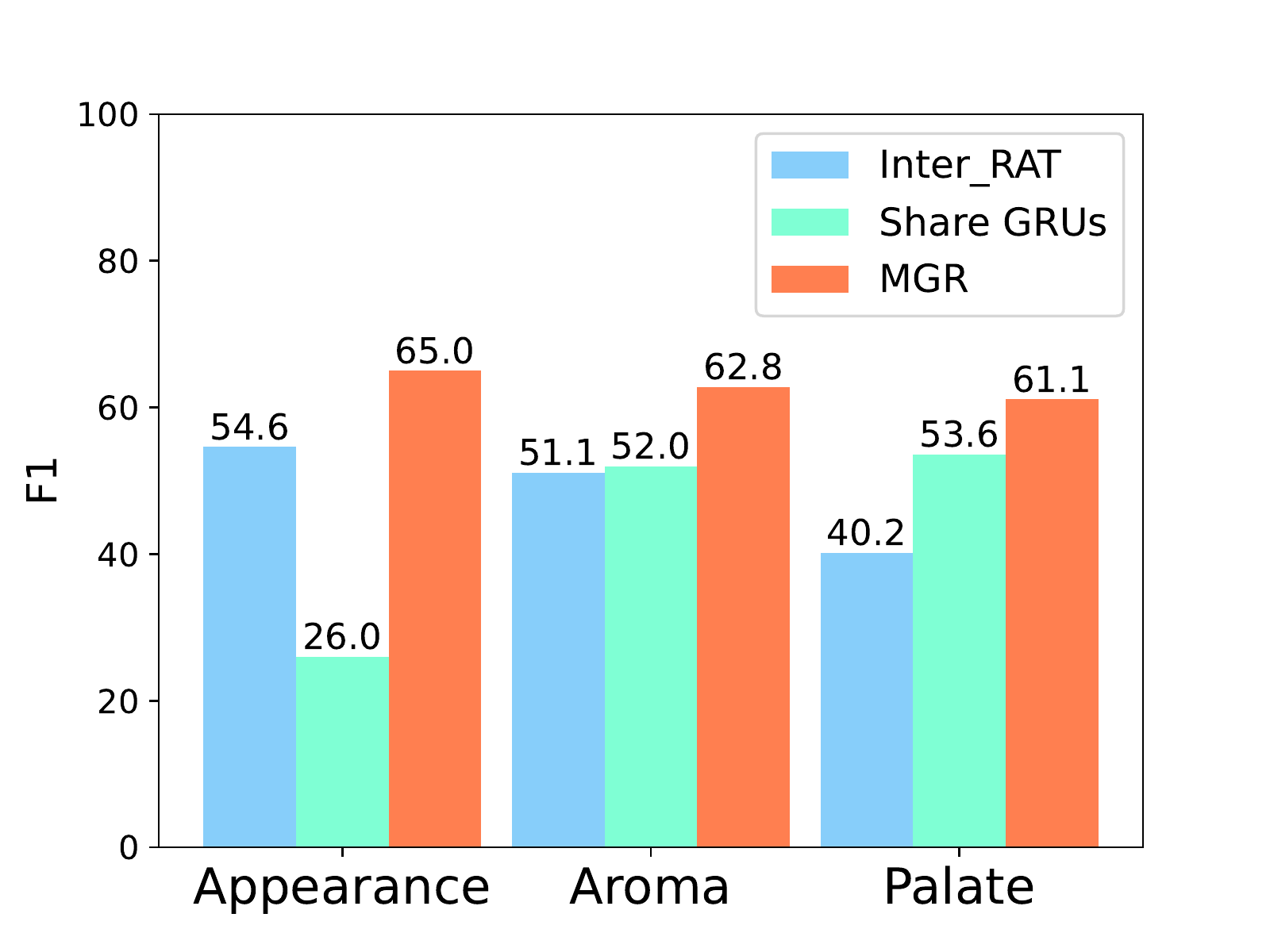}

    \caption{The comparison of sharing the encoders (GRUs) of MGR's generator. The rationale sparsity is around $10\%$.
 }
    \label{fig:share gru sp10}

\end{figure}

\begin{figure}
    \centering

    \includegraphics[width=0.7\columnwidth]{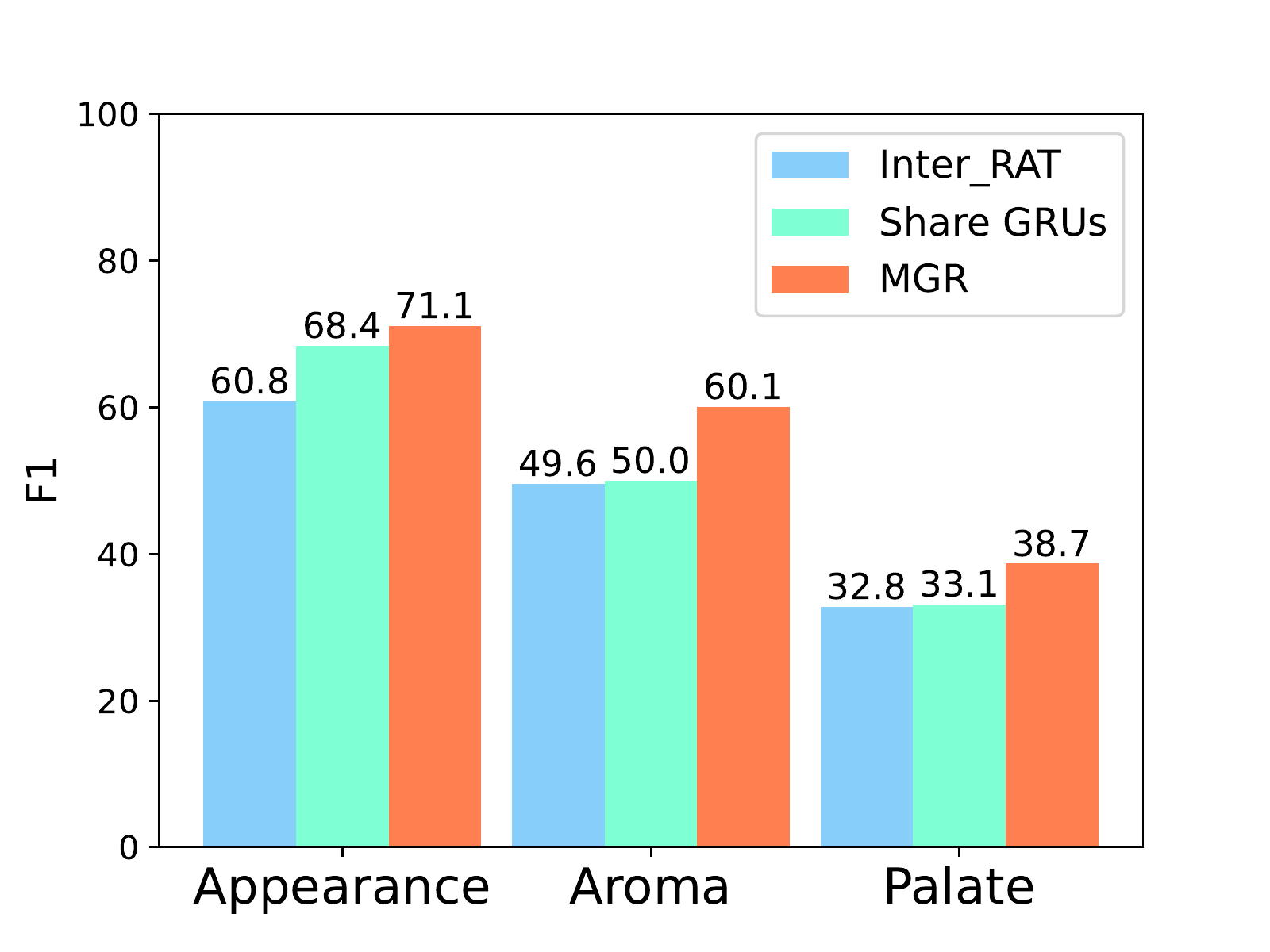}

    \caption{The comparison of sharing the encoders (GRUs) of MGR's generator. The rationale sparsity is around $30\%$.
 }
    \label{fig:share gru sp30}

\end{figure}

 More importantly, how to finetune large models on the rationalization framework is still a significant challenge. Some recent studies \cite{danqi} show that the methods with BERT encoders perform much worse than those with simple GRUs on BeerAdvocate and HotelReviews, which is shown in Table~\ref{tab:res_bert}. VIB and SPECTRA are two RNP-based model. When using BERT, these two methods perform much worse than the vanilla RNP with GRUs (as compared to the results in Table~\ref{tab:beer}).

\subsection{The details of Beer-Skewed}\label{app:skewpred}
The experiment was first designed by \citet{interlocking}. It deliberately induces degeneration to show the robustness of A2R compared to RNP. The predictor is first pre-trained using the first sentence of each text for a few epochs. In Beer Reviews, the first sentence is usually about appearance. So, the predictor will overfit to the aspect of \emph{Appearance}, which is uninformative for \emph{Aroma} and \emph{Palate}. In fact, as compared to degeneration, we think it's more like spurious correlation, which may explain why A2R also fails in this situation.

\subsection{More Results about Sharing the GRUs}\label{app:share grus}
Figure~\ref{fig:share gru sp20} in  the main paper has shown the results of sharing the generators' encoders with the rationale sparsity being around $20\%$, and we further show  the results with the sparsity being around $10\%$ and $30\%$ in Figure~\ref{fig:share gru sp10} and~\ref{fig:share gru sp30}, respectively. Simply sharing the encoders may not be the best way to reduce the computational costs due to the damage on the model performance, but it still outperform Inter\_RAT in most cases. For \emph{Appearance} with $10\%$ sparsity, the reason for the poor performance may come from two aspects. First, as compared to the percentage of human-annotated rationales ($18.4\%$), $10\%$ is too small. It is hard to find the true rationales under such sparsity constraint. Second, the shared encoder limits the explore power of MGR, making the above problem more severe. We will look for better method to reduce the computational costs in the future.

\subsection{The rationale-overlap between different generators}\label{app:rationale difference}
\begin{figure*}[t]
    \centering
    \subfigure[$\frac{||M_1-M_2||_1}{||M_1||_1+||M_2||_1}$]{
        \includegraphics[height=3.3cm]{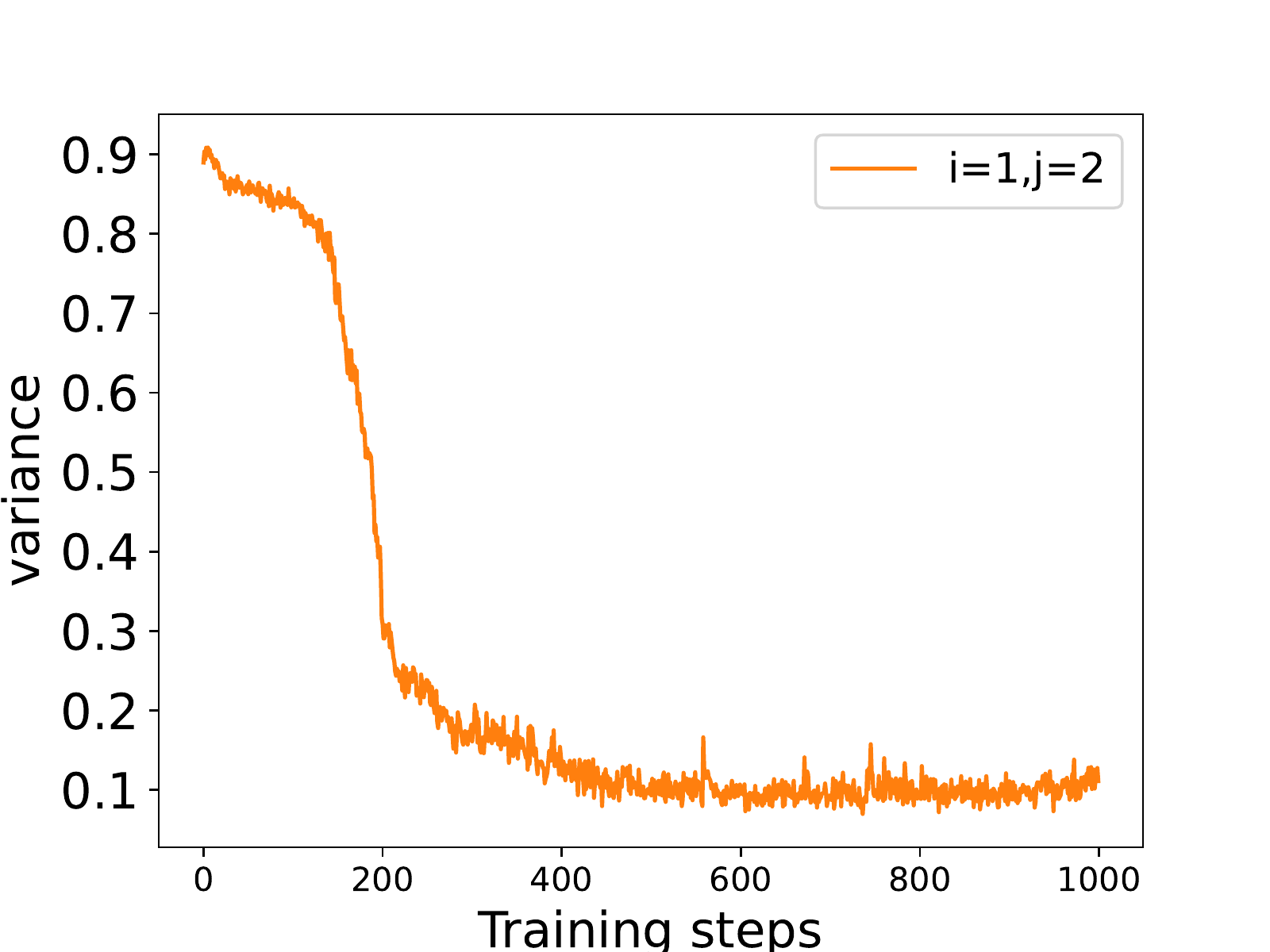}  
    }
   \subfigure[$\frac{||M_1-M_3||_1}{||M_1||_1+||M_3||_1}$]{
        \includegraphics[height=3.3cm]{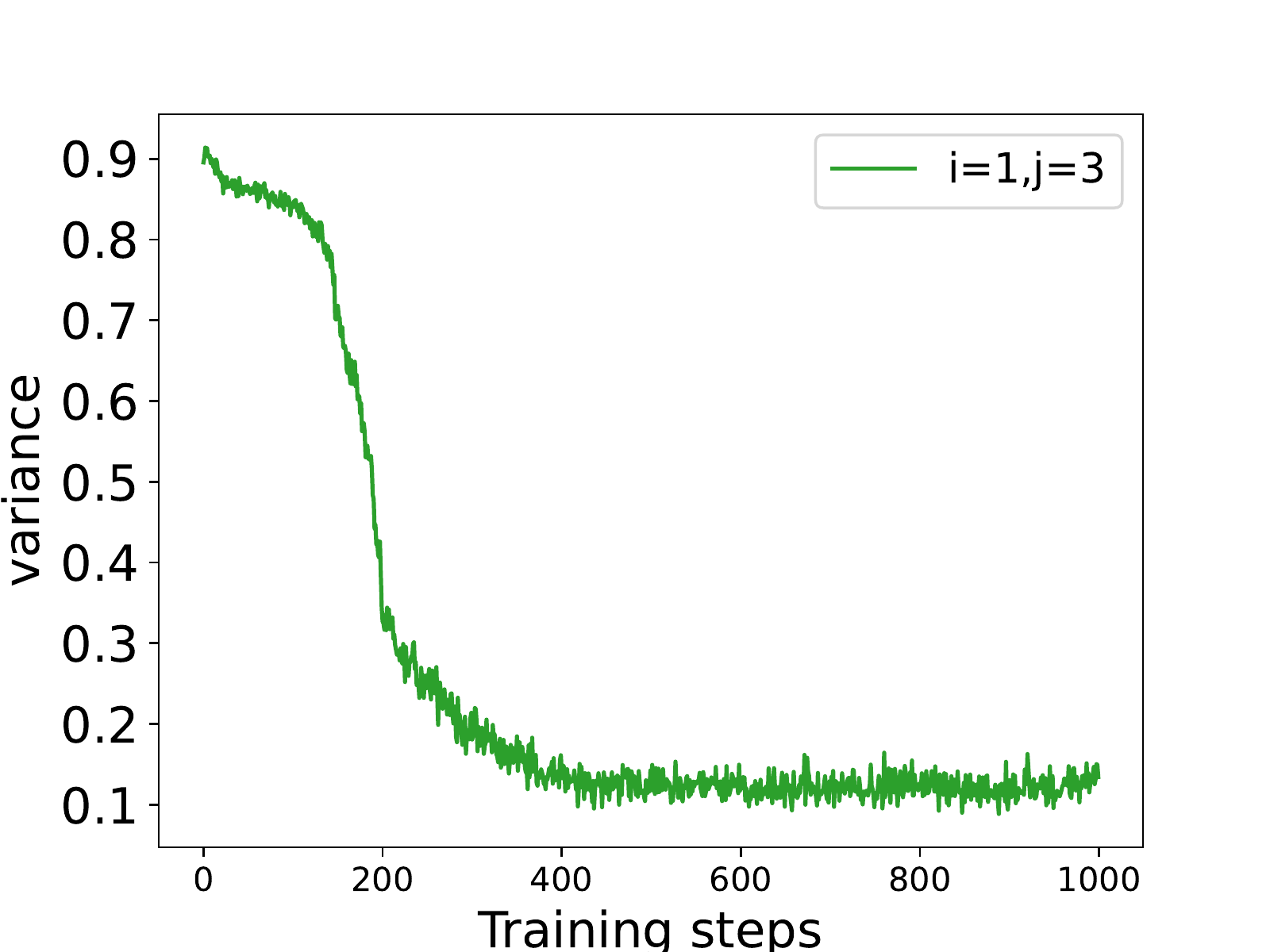}  
    }
    \subfigure[$\frac{||M_2-M_3||_1}{||M_2||_1+||M_3||_1}$]{
        \includegraphics[height=3.3cm]{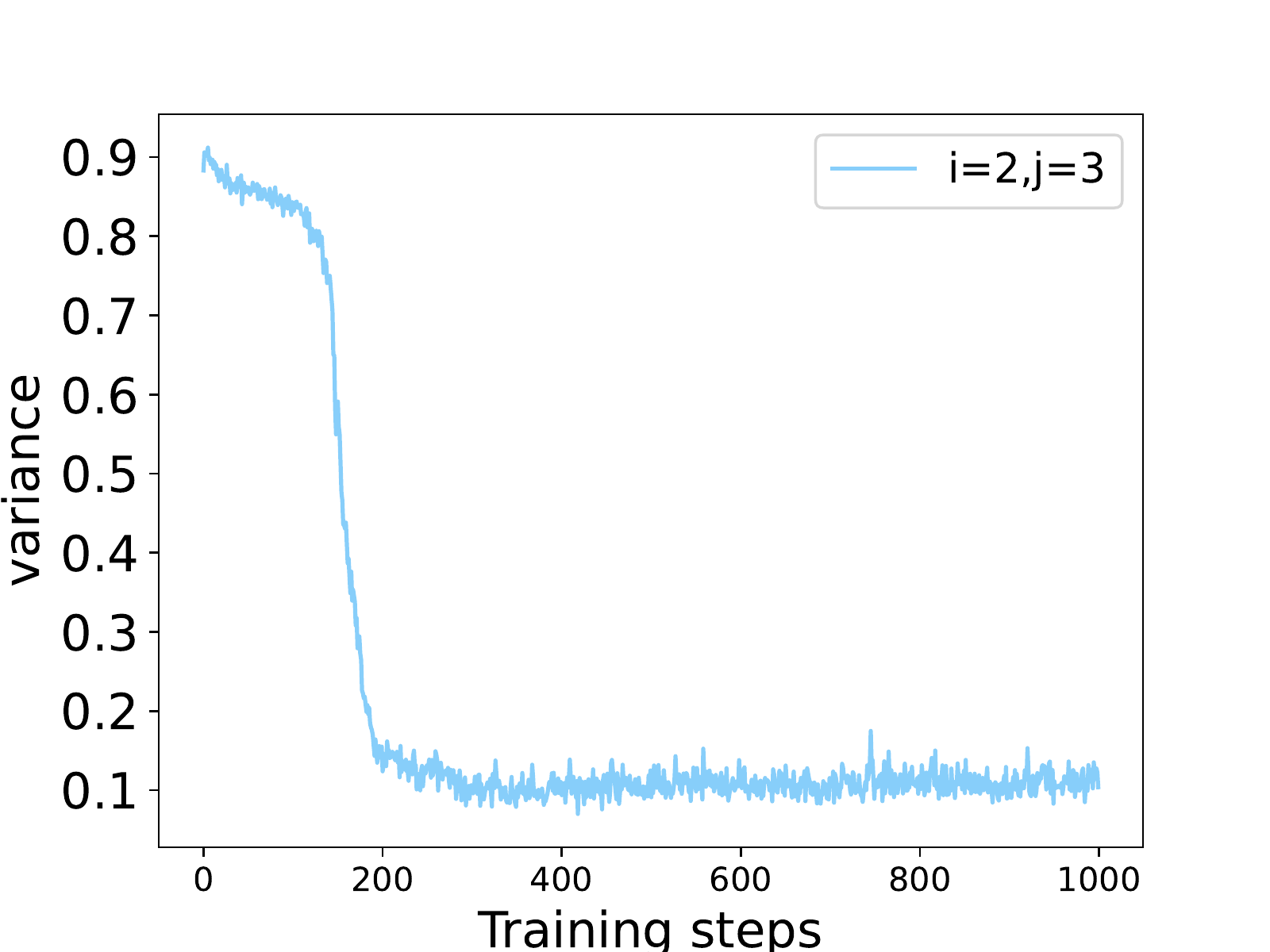}  
    }
    \subfigure[$\frac{||M_1-M_2||_1}{||M_1||_1+||M_2||_1}$]{
        \includegraphics[height=3.3cm]{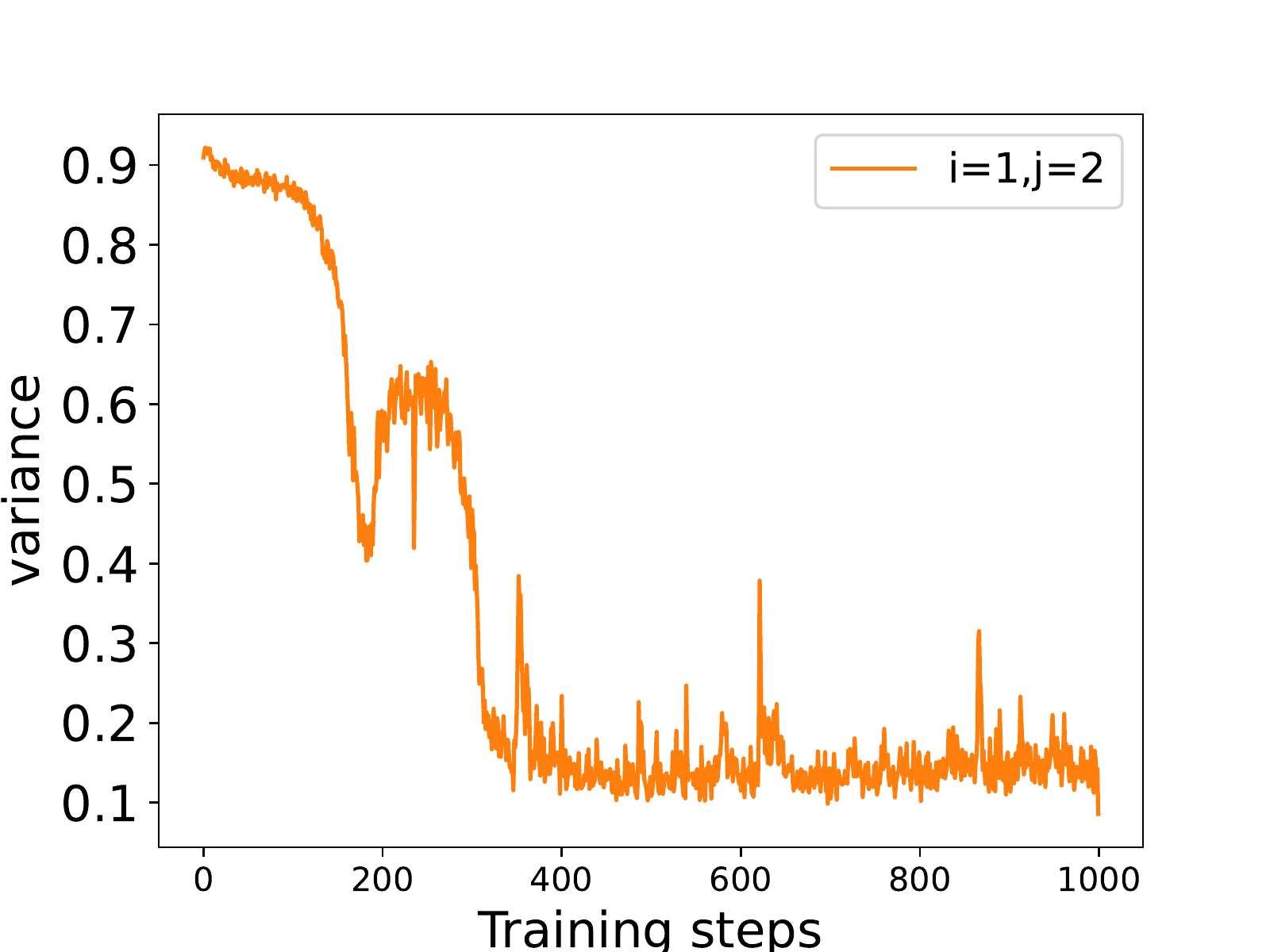}  
    }
   \subfigure[$\frac{||M_1-M_3||_1}{||M_1||_1+||M_3||_1}$]{
        \includegraphics[height=3.3cm]{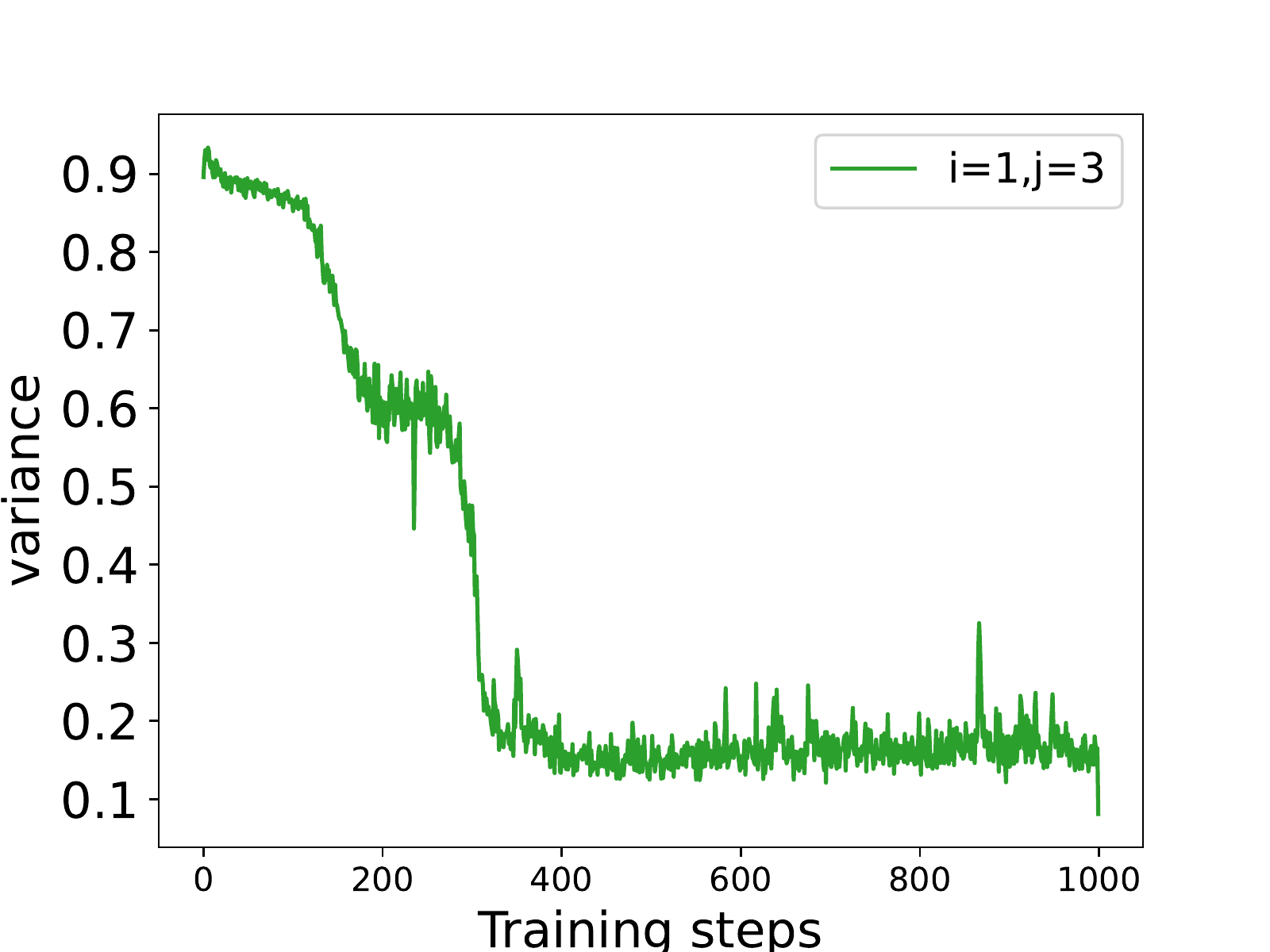}  
    }
    \subfigure[$\frac{||M_2-M_3||_1}{||M_2||_1+||M_3||_1}$]{
        \includegraphics[height=3.3cm]{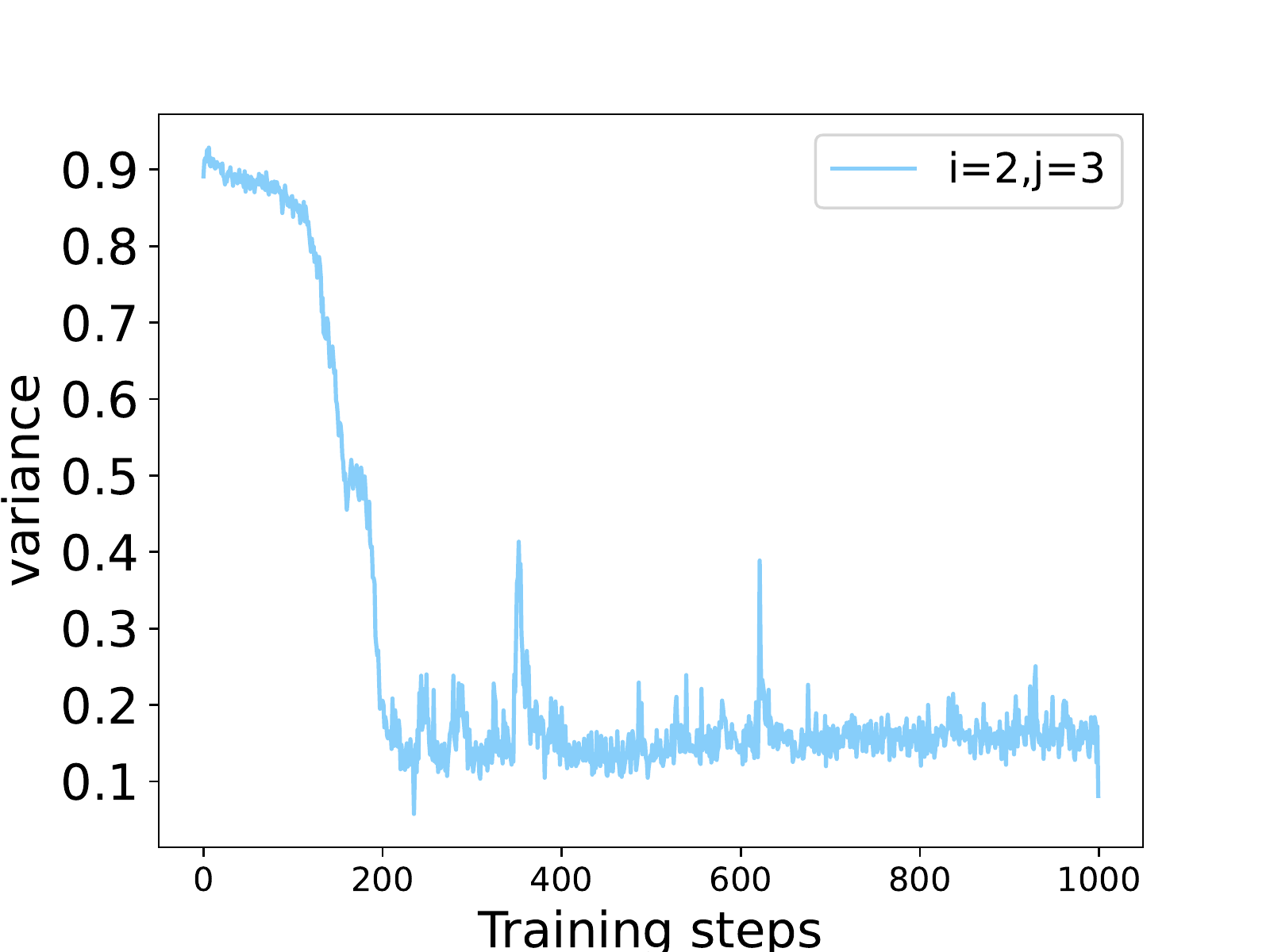}  
    }
    \subfigure[$\frac{||M_1-M_2||_1}{||M_1||_1+||M_2||_1}$]{
        \includegraphics[height=3.3cm]{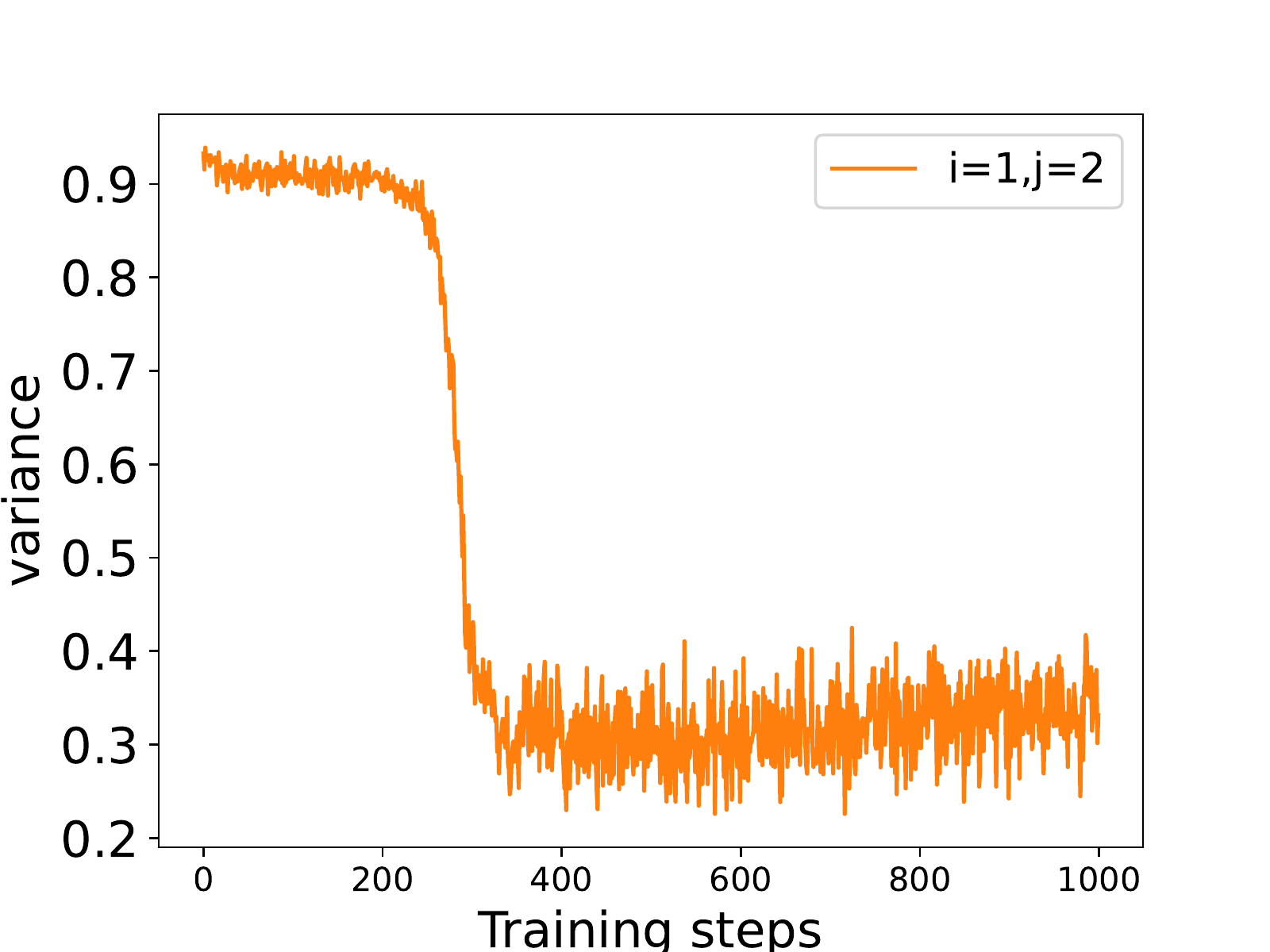}  
    }
   \subfigure[$\frac{||M_1-M_3||_1}{||M_1||_1+||M_3||_1}$]{
        \includegraphics[height=3.3cm]{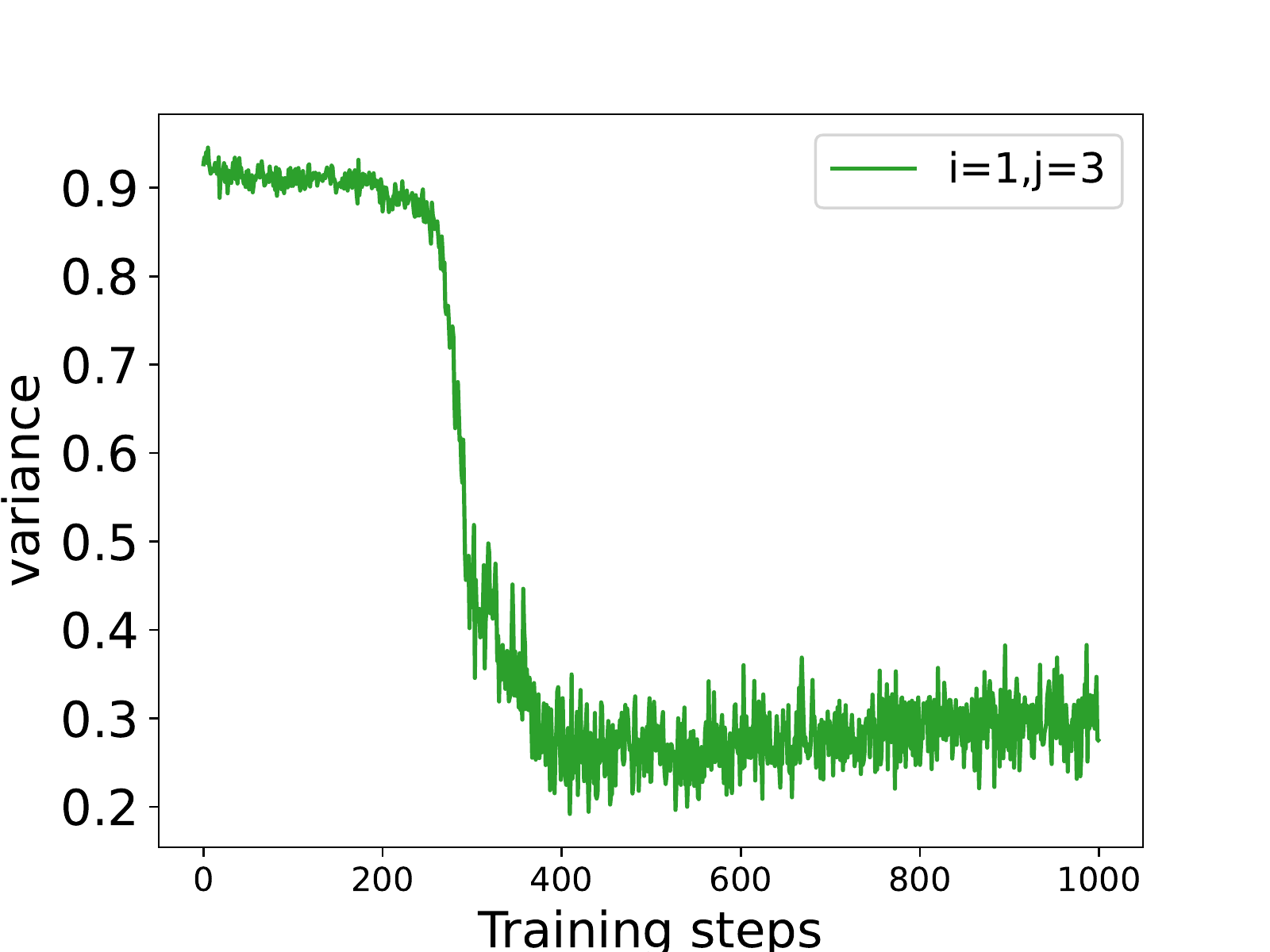}  
    }
    \subfigure[$\frac{||M_2-M_3||_1}{||M_2||_1+||M_3||_1}$]{
        \includegraphics[height=3.3cm]{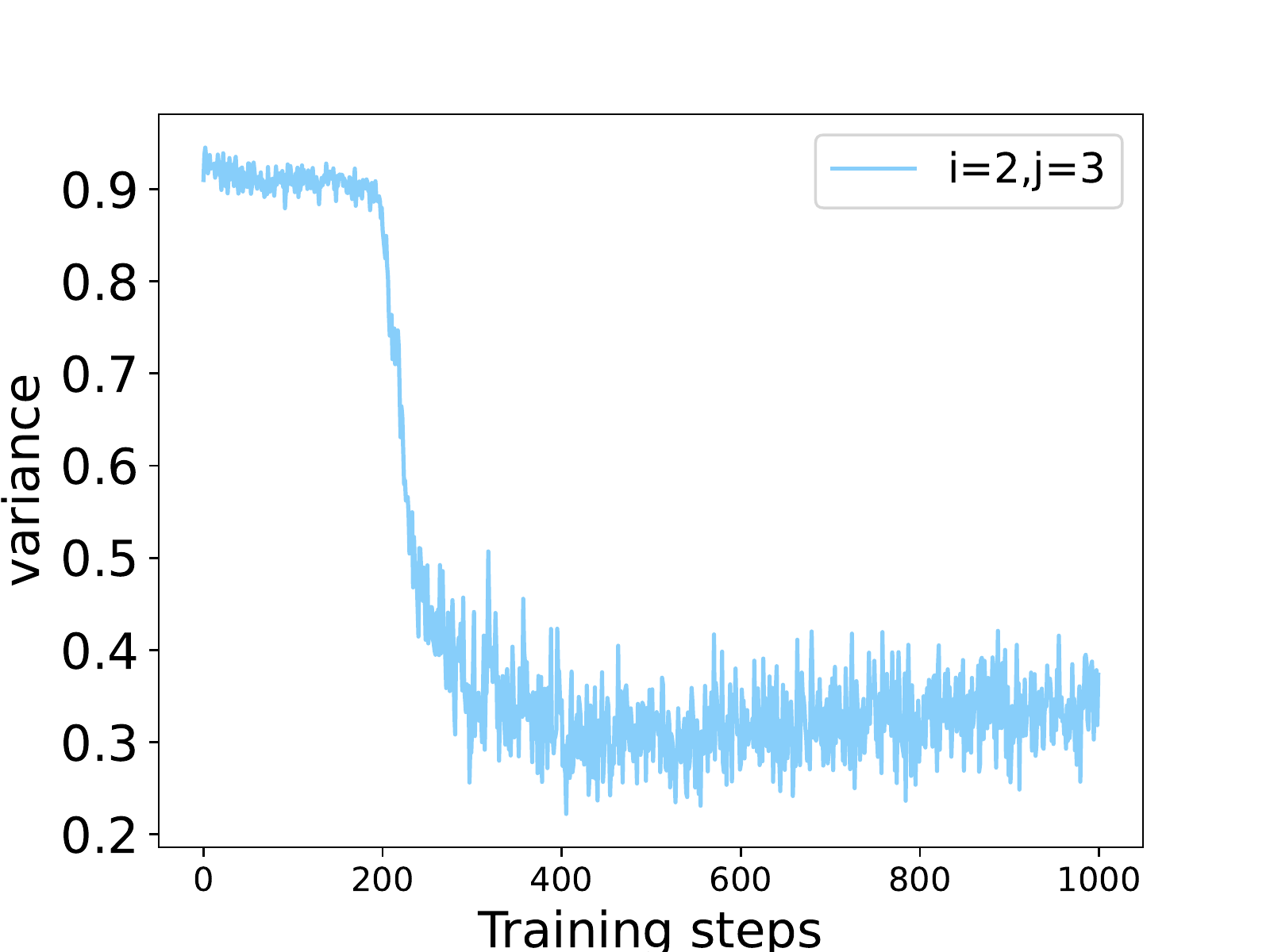}  
    }
  \caption{Rationale overlap from multiple generators on \emph{decorrelated BeerAdvocate}. (a)(b)(c): Appearance aspect. (d)(e)(f): Aroma aspect. (h)(i)(j):Palate aspect. $M_i,M_j$: rationale masks (Equation~\ref{eqa:getrat}) from the $i$-th and $j-th$ generators. \textbf{The y-axis shows the percentage of different tokens} in rationales from different generators.
  }
  \label{fig:difference of rationales from different generators}
\end{figure*}

Corresponding to Figure~\ref{fig:nodis and onehead}(b), we plot the rationale-overlap between different generators in Figure~\ref{fig:difference of rationales from different generators}. The metric is $\frac{||M_i-M_j||_1}{||M_i||_1+||M_j||_1}$, which represents the percentage of different tokens in rationales from different generators. $M_i$ represents the binary rationale mask from the $i$-th generator. The figures show that the variance is high initially and gradually converges to a small value. So, the generators are diverse initially and finally converge to be the same.

\section{Proofs of Theorems}\label{app:proof}

\subsection{Derivation of Equation~\ref{eqa:limit error}}\label{proof:limit error}
To make the presentation succinct, we first discuss the case where $n$ is an odd number.
\begin{equation}
\begin{aligned}
      &\lim_{n\to\infty}p(k<n-k)\\
      =& \lim_{n\to\infty}p(k<\frac{n}{2})   \\
     =& \lim_{n\to\infty}\sum_{k=0}^{(n-1)/2}(C_{n}^{k}\cdot P_c^k\cdot(1-P_c)^{n-k})
\end{aligned}
\end{equation}
Since $P_c>0.5$, we then have 
\begin{equation}
    \lim_{n\to\infty}\sum_{k=0}^{(n-1)/2}(C_{n}^{k}\cdot P_c^k\cdot(1-P_c)^{n-k})=0.
\end{equation}
There is nothing different expect that the upper limit of the summation should be replaced by $n/2-1$ when n is an even number.

\subsection{Proof of Lemma~\ref{lem: pred choose x1}}\label{proof:lemma1}
According to Equation~\ref{eqa:payoff predictor}, we have 
\begin{equation}
\begin{aligned}
    \frac{\partial R_P({\alpha})}{\partial\alpha}= & a\cdot k-(n-k)a-k\cdot b+(n-k)b\\
    = & a\cdot(k-(n-k))-b\cdot(k-(n-k))\\
    = & (a-b)(k-(n-k)).
\end{aligned}
\end{equation}
Since we have $a>b$ and $k>n-k$, we get $\frac{\partial R_P({\alpha})}{\partial\alpha}>0$. It means that to get a higher payoff, the predictor needs to increase $\alpha$, i.e., it needs to move towards $\theta_P^1$. The proof of Lemma~\ref{lem: pred choose x1} is completed.

\subsection{Proof of Theorem~\ref{theorem: n and spurious}}\label{proof: n and spurious}
The proof is obvious. It's equal to that $\lim_{n\to\infty}p_{MGR}(spu)=0$. The left derivation is the same as Appendix~\ref{proof:limit error}.

\subsection{Discussion about $P_c>0.5$}\label{proof:pc>0.5}

\begin{table}[t]
    \centering
    \begin{tabular}{c|c|c|c}
    \hline
      Full   & Decorrelated & Correlated&$P_c$\\
      \hline
         30564&15169&15395&0.67\\

         \hline
    \end{tabular}
    \caption{The $P_c$ approximated by the statistical data of the \emph{Beer-Aroma} dataset. It is approximated by $1-\frac{Correlated}{Correlated*2+Decorrelated}$. We only count samples with negative labels because because the original dataset is unbalanced and we do sampling balance according to the number of negative samples during training.  }
    \label{tab:pc}
\end{table}
For a dataset, there are some samples that contain both the causality and the spurious correlation (i.e., $X_1$ and $X_2$, corresponding to the number of \emph{Correlated} in Table~\ref{tab:pc}), and the other samples contain only the causality (i.e., $X_1$, corresponding to the number of \emph{Decorrelated} in Table~\ref{tab:pc} ). So we always have the number of $X_1$ is larger than that of $X_2$. And for random selection, the probability of selecting $X_1$ is higher than selecting $X_2$, which means that $P_c>0.5$. In Table~\ref{tab:pc}, we approximate $P_c$ by 
\begin{equation}
\begin{aligned}
    P_c=&\frac{\text{Number}(X_1)}{\text{Number}(X_1)+\text{Number}(X_2)}\\
    =&\frac{Decorrelated+Correlated}{Decorrelated+2*Correlated}>0.5\\
    =&1-\frac{Correlated}{Correlated*2+Decorrelated}>0.5.
\end{aligned}
\end{equation}

\subsection{Proof of Theorem~\ref{theorem: higher entropy with mgr}}\label{proof:higher entropy of mgr}
We first proof the left inequality of Theorem~\ref{theorem: higher entropy with mgr}.
For any two random variable ${Z_i},{Z_j}$, we have 
\begin{equation}
\begin{aligned}
       H(Z_i|Z_j)&=H(Z_i,Z_j)-H(Z_j) \\
       H(Z_i|Z_j)&\geq 0\\
       H(Z_j)&\geq 0.
\end{aligned}
\end{equation}
So, we have 
\begin{equation}
    H(Z_i,Z_j)\geq H(Z_j),
\end{equation}
where the equality holds if and only if $H(Z_i|Z_j)=0$, i.e., $Z_i=Z_j$.
There is nothing different for $H(Z_i,Z_j)\geq H(Z_i)$. Then we easily get the left inequality of Theorem~\ref{theorem: higher entropy with mgr} through Mathematical Induction.

Then we proof the right inequality of Theorem~\ref{theorem: higher entropy with mgr}. We first have 
\begin{equation}
\begin{aligned}
       I(Z_i,Z_j)&=H(Z_i)+H(Z_j)-H(Z_i,Z_j)\\
       I(Z_i,Z_j)&\geq 0,
\end{aligned}
\end{equation}
where $I(Z_i,Z_j)$ is the mutual information. $I(Z_i,Z_j)=0$ if and only if $ Z_i \upmodels Z_j$.
So, we have 
\begin{equation}
    H(Z_i,Z_j)\leq H(Z_i)+H(Z_j),
\end{equation}
with the equality holds if and only if  $ Z_i \upmodels Z_j$.
Then we easily get the right inequality of Theorem~\ref{theorem: higher entropy with mgr} through Mathematical Induction.

The proof of Theorem~\ref{theorem: higher entropy with mgr} is completed.

\clearpage

\end{document}